\documentclass[journal]{IEEEtran}
\usepackage[numbers]{natbib}
\usepackage{graphicx}
\usepackage{multirow}
\usepackage{epstopdf}
\usepackage{times}
\usepackage{amsmath}
\usepackage{amssymb}
\usepackage{amsfonts}
\usepackage{color}
\usepackage{subfigure}
\usepackage{xcolor}
\usepackage{algorithm}
\usepackage{algorithmic}
\usepackage{rotating}
\usepackage{adjustbox,lipsum}
\usepackage{dsfont}

\usepackage{booktabs}
\usepackage{multirow}
\usepackage{array}
\usepackage{graphicx}
\usepackage{tabularx}
\usepackage{colortbl} 
\usepackage{soul}
\definecolor{darkgreen}{HTML}{05c880}
\usepackage[colorlinks=true, citecolor=darkgreen, linkcolor=purple]{hyperref}
\definecolor{yellow_color}{HTML}{fde724}
\definecolor{purple_color}{HTML}{AF58BA}
\definecolor{blue_color}{HTML}{009ADE}
\definecolor{morandigreen}{HTML}{9DD3A8}
\definecolor{morandired}{HTML}{D9827E}
\definecolor{morandiblue}{HTML}{A3B5C6}
\definecolor{tcsvtblue}{rgb}{0.21,0.49,0.74}

\newcommand{\ie}{{\em i.e.}}           

\definecolor{lightgreen}{RGB}{97, 151, 151}
\begin{document}

\title{A Novel Cross-Perturbation for Single Domain Generalization}

\author{Dongjia Zhao,
        Lei Qi,
        Xiao Shi,
        Yinghuan Shi,
        Xin Geng

\thanks{The work is supported by NSFC Program (Grants No. 62206052, 62125602, 62076063), Jiangsu Natural Science Foundation Project (Grant No. BK20210224), and the Xplorer Prize.}
\thanks{Dongjia Zhao, Lei Qi, Xiao Shi and Xin Geng are with the School of Computer Science and Engineering, Southeast University, and Key Laboratory of New Generation Artificial Intelligence Technology and Its Interdisciplinary Applications (Southeast University), Ministry of Education, China, 211189 (e-mail: zhaodongjia@seu.edu.cn; qilei@seu.edu.cn; xshi@seu.edu.cn; xgeng@seu.edu.cn).}
\thanks{Yinghuan Shi is with the State Key Laboratory for Novel Software Technology, Nanjing University, Nanjing, China, 210023 (e-mail: syh@nju.edu.cn).}
\thanks{Corresponding author: Lei Qi.}

}

%
%

\markboth{~}%
{Shell \MakeLowercase{\textit{et al.}}: Bare Demo of IEEEtran.cls for IEEE Journals}

\maketitle

\begin{abstract}
Single domain generalization aims to enhance the ability of the model to generalize to unknown domains when trained on a single source domain. However, the limited diversity in the training data hampers the learning of domain-invariant features, resulting in compromised generalization performance. To address this, data perturbation (augmentation) has emerged as a crucial method to increase data diversity. Nevertheless, existing perturbation methods often focus on either image-level or feature-level perturbations independently, neglecting their synergistic effects. To overcome these limitations, we propose CPerb, a simple yet effective cross-perturbation method. Specifically, CPerb utilizes both horizontal and vertical operations. Horizontally, it applies image-level and feature-level perturbations to enhance the diversity of the training data, mitigating the issue of limited diversity in single-source domains. Vertically, it introduces multi-route perturbation to learn domain-invariant features from different perspectives of samples with the same semantic category, thereby enhancing the generalization capability of the model. Additionally, we propose MixPatch, a novel feature-level perturbation method that exploits local image style information to further diversify the training data. Extensive experiments on various benchmark datasets validate the effectiveness of our method.
\end{abstract}


%
\IEEEpeerreviewmaketitle

\section{Introduction}
\IEEEPARstart{I}{n} recent years, domain generalization~\cite{choi2021robustnet, huang2020self} has become a crucial and pressing challenge in the transfer learning field~\cite{jia2019semi, meng2022exploring}, especially in single-source scenarios. This challenge stems from the unavoidable distribution disparities between the source (training) and target (testing) domains, leading to a notable decline in generalization performance when deep learning networks are trained on a specific source domain and tested on previously unseen target domains. This limitation hinders the practical deployment of deep learning models in real-world scenarios due to the inevitable distribution shift between training and testing datasets.

\begin{figure}[!h]
\centering
\subfigure[Mean 
 and Variance of \underline{means}]{
\begin{minipage}[b]{0.92\linewidth}
\centering
\includegraphics[width=\textwidth]{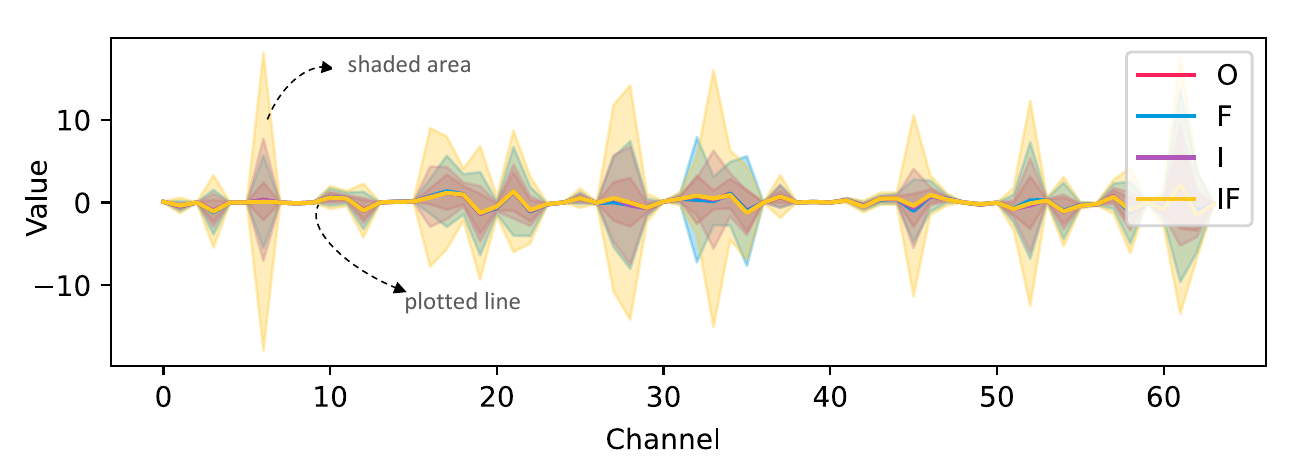}
\end{minipage}
}
\subfigure[Mean and Variance of \underline{variances}]{
\begin{minipage}[b]{0.92\linewidth}
\centering
\includegraphics[width=\textwidth]{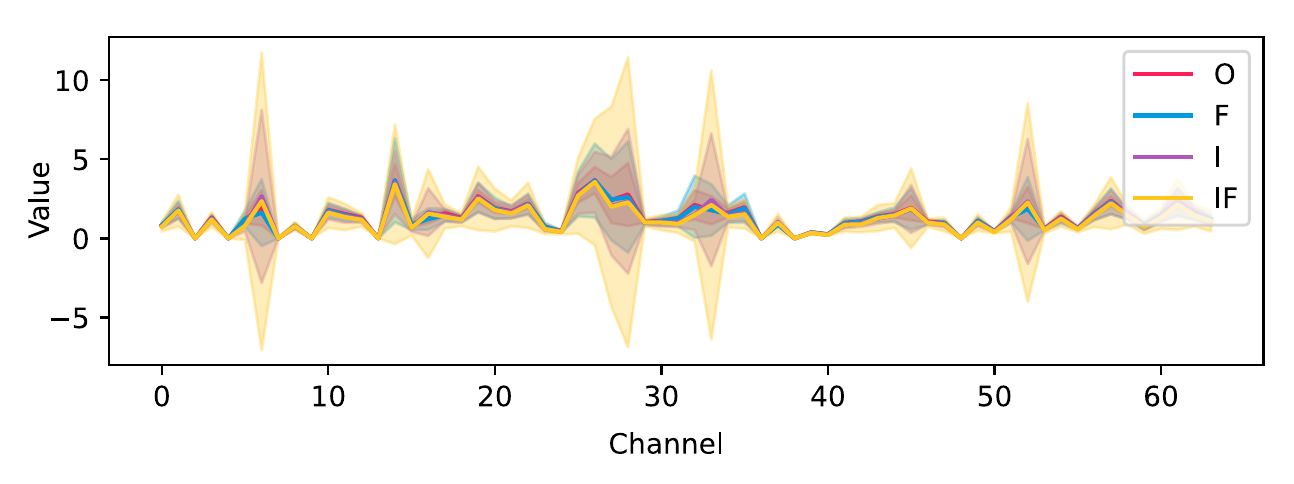}
\end{minipage}
}
\caption{Statistical discrepancies among different perturbation methods. Experiment on 128 images from ``art\_painting'' domain on PACS. ``O'' and ``I'' refer to the original image and image-level perturbation, respectively, both processed through the first convolutional layer of ResNet18 pre-trained on ImageNet to obtain output feature maps. ``F'' and ``IF'' indicate the original image and image-level perturbation, respectively, where the feature-level perturbation is applied after extracting features from the convolutional layer. Mean and variance (\ie, statistics) are computed for the feature on each original image (O), image-level perturbation (I), feature-level perturbation (F), or image-feature dual-level perturbation (IF). Thus, for each channel, we can obtain 128 means and variances. Then, we compute the mean (\ie, the plotted lines) and variance (\ie, the shaded areas) of 128 means, as shown in (a). Similar to the mean and variance of 128 variances in (b).}
\label{fig-cperb-horizontal-motivation}
\vspace*{-18pt}
\end{figure}

To mitigate the impact of domain shift, a multitude of domain generalization techniques have emerged~\cite{zhang2022exact, zhang2022mvdg}. Many of these methods have been developed for the context of multi-source domain generalization, wherein multiple domains are collectively utilized during training. This strategy benefits from the abundance of training samples and the ability to simulate domain shift by leveraging the discrepancies among source domains. As a result, these methods have demonstrated remarkable advancements in multi-source domain generalization. In the realm of single-source domain generalization~\cite{du2020learning}, the limited diversity of samples presents a significant obstacle to acquiring domain-invariant features. Consequently, data perturbation has emerged as a viable strategy for augmenting data diversity. In this study, our main focus is to tackle the specific challenge of single-source domain generalization.

Numerous prior studies have made noteworthy contributions to tackling the challenge of single-source domain generalization. These methods utilize a range of techniques, including adversarial training~\cite{fan2021adversarially, chen2023center, li2021triple, zhang2024multi} and data augmentation~\cite{wang2022feature, wang2021learning, qiao2020learning, aversa2020deep}, to mitigate the effects of domain shift to some extent. Their main goal is to improve the generalization ability of the network by reducing the domain shift between the source and target domains. For example, ASR-Norm~\cite{ASR} introduces adaptive normalization and rescaling components into the normalization process. DSU~\cite{dsu} estimates the variance of feature statistics to generate feature statistical variables that encompass various distribution possibilities, replacing the original deterministic values and thereby simulating diverse domain shifts. EFDMix~\cite{zhang2022exact} achieves precise feature distribution matching in the feature space using higher-order statistics and augments the training data with style transfer techniques to mitigate overfitting to the source domain.

Nevertheless, despite the noteworthy accomplishments of prior methods, numerous challenges and limitations persist. A significant limitation lies in the fact that existing techniques, which aim to enhance model robustness through data augmentation, often prioritize either image-level or feature-level perturbations, neglecting the complementary interplay between perturbations at different levels. Furthermore, existing instance-level feature perturbation methods fail to exploit the inherent variations in the internal style information of images.


Recently, research has shown that the style information of an image can be effectively represented using its mean and variance (\ie, statistics)~\cite{huang2017arbitrary}. As depicted in Fig.~\ref{fig-cperb-horizontal-motivation}, our observations reveal that the combination of image-level and feature-level perturbations results in the broadest range of statistical data variability (indicated by the \setulcolor{yellow_color}\ul{yellow} region), effectively enhancing the diversity of the data. Additionally, notable style distinctions exist among image-level perturbation, feature-level perturbation, and image-feature dual-level perturbation. The incorporation of multiple perspectives of the same image aids the network in extracting domain-invariant features.

Motivated by these valuable insights, we propose a novel and effective cross-perturbation method named CPerb. Our method aims to enhance the robustness and generalization ability of deep learning models by providing complementary data perturbation at two levels. Specifically, the method encompasses both horizontal and vertical operations. In the horizontal operations, we joint leverage image- and feature-level perturbations to enrich the diversity of the training data, addressing the issue of limited diversity in single-source domains. In the vertical operations, we introduce a multi-route augmentation method that facilitates the learning of domain-invariant features and improves  generalization performance of the model by considering multiple perspectives of samples with the same semantic category.

Additionally, to further enhance the feature-level perturbations, we propose a novel method called MixPatch. This method exploits the differences in local image style information to augment the diversity of the training data. It achieves this by expanding and exchanging local style statistics between images, thereby reducing the domain discrepancy between the source and target domains. Consequently, our proposed method, CPerb, outperforms state-of-the-art (SOTA) methods across multiple tasks. In summary, the main contributions of this paper are as follows:
 
\begin{itemize}
 
\item[$\bullet$] We propose CPerb, a cross-perturbation method that enhances data diversity through horizontal operations involving image- and feature-level perturbations, and promotes the learning of domain-invariant features through vertical operations using multi-route perturbation. 
\item[$\bullet$] In order to further enhance the diversity of training data by effectively incorporating local image style information, we introduce MixPatch, a novel method that introduces perturbations at the feature level, thereby enriching the training data and improving the ability of the model to handle style variations.  
\item[$\bullet$] We conduct extensive experiments on various benchmark for image classification and instance retrieval to evaluate the performance of the proposed CPerb framework. By comparing it against state-of-the-art methods, we demonstrate the effectiveness and superiority of our method in achieving improved robustness across different domains.

\end{itemize}

The rest of this paper is organized as follows.
The related work is reviewed in Section \ref{s-related}.
The CPerb framework is proposed and discussed in Section \ref{s-framework}.
Experimental results and analysis are presented in Section \ref{s-experiment},
and the conclusion is drawn in Section \ref{s-conclusion}.

\section{Related work}\label{s-related}

\subsection{Domain Generalization}
Due to the inherent distribution disparities between multiple source domains and the target domain, domain generalization~\cite{choi2021robustnet, huang2020self, zhang2022exact, zhang2022mvdg} has emerged as a prominent research area. Distinguishing domain generalization from domain adaptation~\cite{zhao2020multi, sun2015survey, fang2023three, peng2023rain, peng2023source}, the former lacks visibility of target domain data during training, whereas the latter has access to unlabeled target domain data, thereby rendering domain generalization a more arduous task. To tackle these challenges, Ding \textit{et al.}~\cite{ding2022domain} aim to develop a domain-invariant model by strategically eliminating domain-specific features from input images. From the perspective of attention diversification, Meng \textit{et al.}~\cite{meng2022attention} mitigate domain shift issues by incorporating in-channel discrimination and cross-channel diversification techniques on high-level feature maps, thus simulating domain shift and applying regularization. Additionally, meta-learning is employed in the domain generalization task, where Du \textit{et al.}~\cite{du2020learning} introduce a probabilistic meta-learning model and leverages the meta variational information bottleneck principle. The probabilistic meta-learning model represents shared classifier parameters across different domains as distributions, enabling better management of prediction uncertainty in unseen domains. Simultaneously, the meta variational information bottleneck principle establishes novel variational bounds of mutual information, facilitating the acquisition of domain-invariant representations.

\begin{figure*}
\centering
\includegraphics[width=18cm]{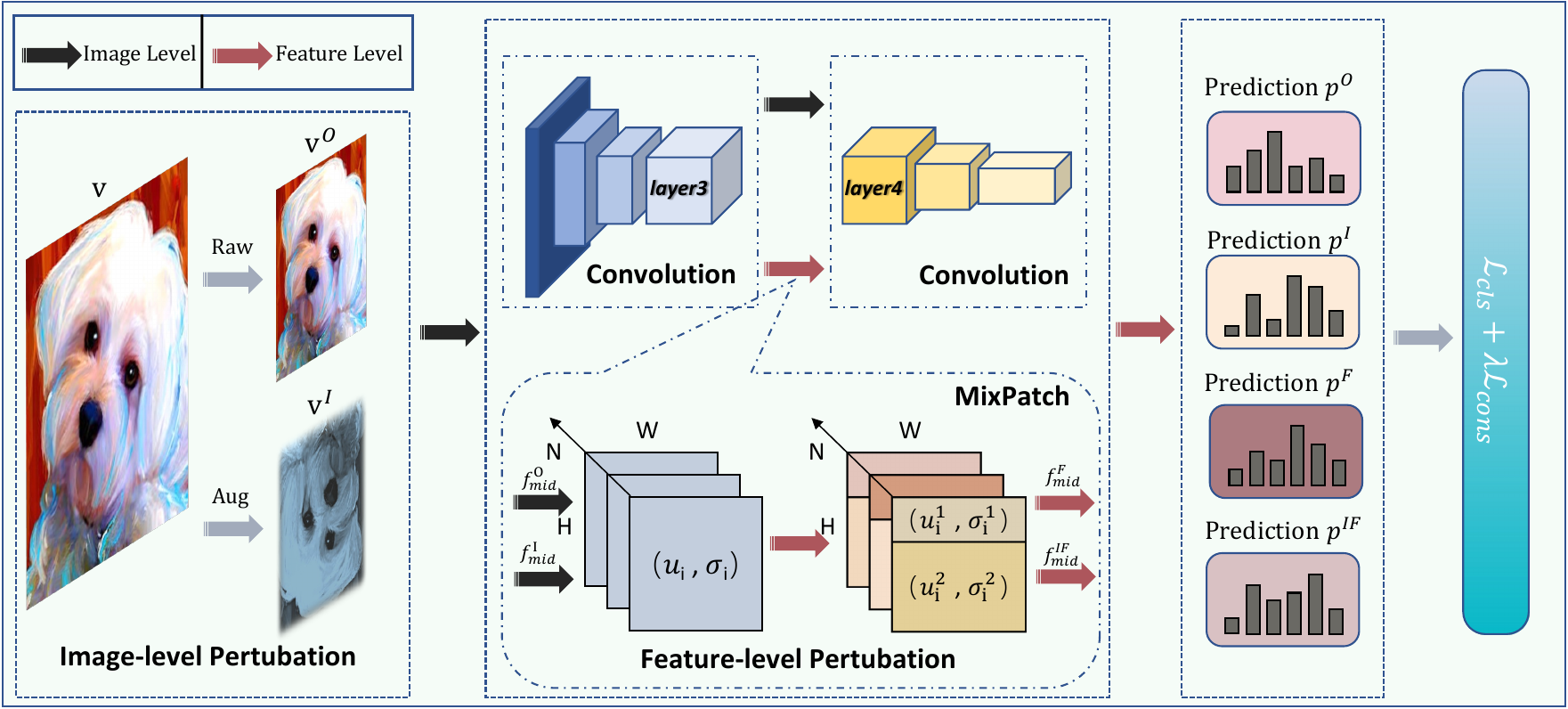}
\caption{Overview of our CPerb framework. The framework involves two paths for the same image: one with image-level augmentation and the other without augmentation, resulting in two images with distinct perspectives ($\text{v}^O$ and $\text{v}^I$). These images are fed into two shared-weight networks, one with feature perturbation and the other without, leading to four predictions used for classification and consistency learning.}
\label{fig3}
\vspace*{-20pt}
\end{figure*}

\subsection{Single Domain Generalization}
Single-source domain generalization~\cite{du2020learning} is a more challenging task due to the lack of diversity in the training set and greater domain shift between the source and target domains. To address this issue, from the perspective of normalization methods~\cite{choi2021robustnet, fan2021adversarially, pan2018two}, the method proposed by Choi \textit{et al.}~\cite{choi2021robustnet} decouple the domain-specific style and domain-invariant content encoded in higher-order statistics of the feature representations and selectively removes style information that only causes domain shift, thereby reducing overfitting. Pan \textit{et al.}~\cite{pan2018two} cleverly integrate instance normalization and batch normalization to extract domain-invariant features, resulting in improved generalization performance. Fan \textit{et al.}~\cite{fan2021adversarially} complement the existing research on normalization layer statistics by learning the statistics of standardization and rescaling via neural networks, adapting to data from different domains and improving model generalization across domains. Wan \textit{et al.}~\cite{wan2022meta} propose a novel meta convolutional neural network that decomposes the convolutional features of images into meta features. By eliminating irrelevant features of local convolutional features through compositional operations and reformulating the convolutional feature maps as a composition of related meta features with reference to meta features, the images are encoded without biased information from unseen domains. Wu \textit{et al.}~\cite{wu2022single} introduce cyclic-disentangled self-distillation for single-source domain object detection, which can disentangle domain-invariant features from domain-specific representations without the need for domain-related annotations, further enhancing the generalization ability of the model.

\subsection{Augmatation for Domain Generalization}
Data augmentation methods are widely adopted in addressing the challenges of single-source domain generalization. These methods can be categorized into image-level-based~\cite{wang2021learning, cugu2022attention} and feature-level-based~\cite{qiao2020learning, zhao2020maximum, volpi2018generalizing, zheng2022faster} methods.

\textbf{Image-level perturbation:} Wang \textit{et al.}~\cite{wang2021learning} focus on generating diverse images by minimizing the upper bound of mutual information between the generated and source samples, while maximizing the mutual information between samples of the same semantic to facilitate the learning of discriminative features. Cugu \textit{et al.}~\cite{cugu2022attention} enhance generalization by introducing visual corruptions to expand the source domain distribution and ensuring consistency between original and augmented samples based on attention consistency.

\textbf{Feature-level perturbation:} In the context of meta-learning, Qiao \textit{et al.}~\cite{qiao2020learning} employ a Wasserstein Auto-Encoder (WAE) to relax the worst-case constraint, enabling effective domain augmentation. Zhao \textit{et al.}~\cite{zhao2020maximum} propose a novel regularization term that perturbs the source data distribution to generate challenging adversarial examples, thereby enhancing the robustness of the model. Volpi \textit{et al.}~\cite{volpi2018generalizing} iteratively augment the dataset with samples from a fictitious target domain, focusing on improving generalization performance.

In contrast to the previously mentioned methods, our method combines both image and feature dual-level perturbations at the horizontal level, with the goal of enriching the diversity of the source domain data. Additionally, at the vertical level, we employ multiple routes of perturbation to effectively learn domain-invariant features.

\section{Method}\label{s-framework}
It is widely acknowledged that single-source domain generalization encounters limitations due to its reliance on a single training dataset, which lacks diversity. To overcome this challenge, data perturbation has emerged as a recognized method to enhance data diversity. However, existing methods often apply image-level or feature-level perturbations in isolation, disregarding the mutually beneficial interaction between these two levels. In order to address this limitation, we propose an effective and straightforward framework named CPerb. In Section \ref{SEC:cperb}, we present a detailed explanation of the CPerb framework, examining both the horizontal and vertical perspectives. In Section \ref{SEC:LDII}, we introduce a novel feature-level perturbation method called MixPatch, which aims to fully exploit the diversity of style information in local image region.



\subsection{Our Holistic Framework: CPerb}\label{SEC:cperb}
As illustrated in Fig.~\ref{fig-cperb-horizontal-motivation}, we have observed that the integration of image-level and feature-level perturbations results in a notable enhancement of data diversity. Additionally, there are distinct stylistic differences among image-level perturbation, feature-level perturbation, and image-feature dual-level perturbation. Drawing inspiration from these insights, we propose a novel cross-perturbation method known as CPerb, encompassing both horizontal and vertical perturbations. Addressing the challenge of limited data diversity in single-source domain generalization training, we delve into the details of the horizontal level. Furthermore, to enhance the generalization capacity of the model through the acquisition of domain-invariant features, we expound upon the vertical level. The comprehensive CPerb framework is visually depicted in Fig.~\ref{fig3}.

\subsubsection{Horizontal-level}\label{SEC:LDI}
Data perturbation plays a crucial role in addressing the problem of insufficient data in single-source domain generalization. However, existing data perturbation methods often focus on either image-level or feature-level perturbations separately, without fully considering the complementary interaction between these two levels. To address this limitation, we propose a two-level perturbation method in the horizontal operations, which incorporates both image-level and feature-level perturbations to enhance the diversity of the source domain distribution. The image-level perturbation in the CPerb is aligned with other perturbation techniques~\cite{french2019semi, zou2020pseudoseg}, which include horizontal random flipping, color transformations, and random grayscale variations. Feature-level perturbations in the CPerb framework utilize the MixPatch, with comprehensive details provided in Section \ref{SEC:LDII}.


\subsubsection{Vertical-level}\label{SEC:LD}
Motivated by recent advancements in self-supervised and semi-supervised learning, the construction of multiple-stream perturbations for image data emerges as a promising method to effectively leverage perturbations for learning domain-invariant features. A good example of this is the UniMatch~\cite{yang2022revisiting}, which uses multi-stream image strong augmentation and strengthens the consistency between two strong augmented flows using the InfoNCE~\cite{oord2018representation} loss.

As depicted in Fig.~\ref{fig-cperb-horizontal-motivation}, we can observe notable differences in feature statistical data between the original image, image-level perturbation, feature-level perturbation, and image-feature dual-level perturbation. This suggests that using different perturbation methods on the same image leads to noticeable variations in feature styles. Hence, adopting a multi-route learning method helps the network in acquiring domain-invariant features. Additionally, to maintain the consistency of domain-invariant features learned across multiple routes, we introduce the concept of a consistency loss.

\begin{figure}
\centering
\includegraphics[width=8.5cm]{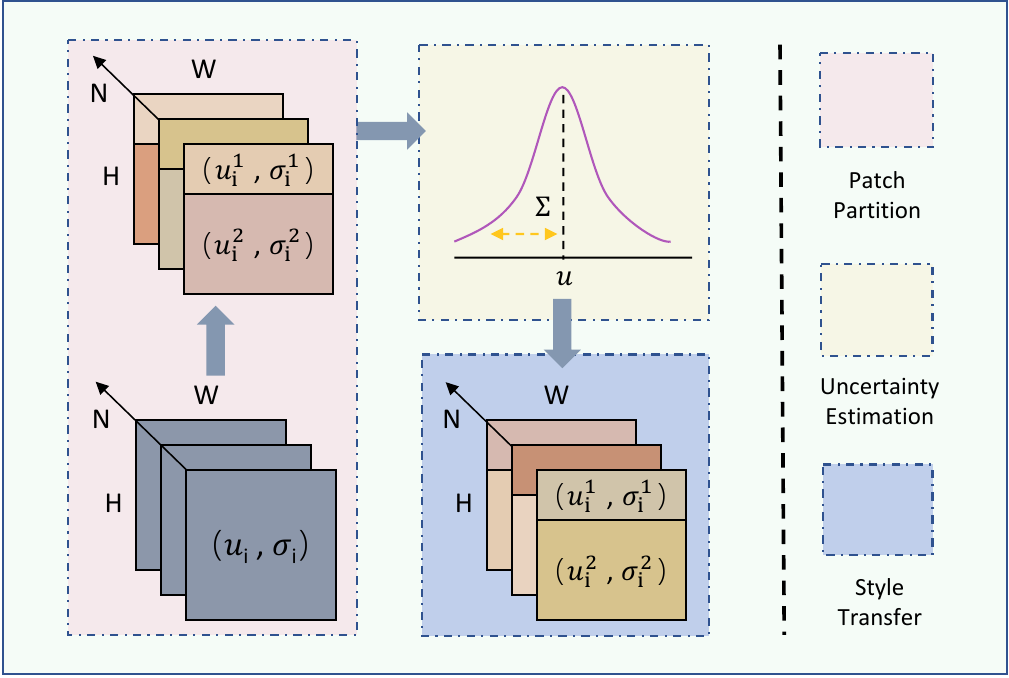}
\caption{Illustration of the MixPatch framework. Initially, the feature maps are partitioned into patches along the channel dimension. Subsequently, uncertainty estimation is conducted at the patch level within each channel to uncover potential style variations. Finally, style transfer is employed on the rescaled patches within each channel, leading to the generation of the fused feature maps.}
\label{fig-MixPatch}
\vspace*{-15pt}
\end{figure}

In our method, we obtain four outputs, namely $p_i^O$, $p_i^I$, $p_i^F$, and $p_i^{IF}$. To ensure consistency among the multiple routes, we use Jensen-Shannon divergence (JS) to measure the differences between the predictions, $p_{i}^{perb}$, with $p_i^{mean}$ representing the average predicted value across all routes.

\begin{equation}\label{eq15}
  \begin{aligned}
    \begin{gathered}
      p_{i}^{mean}=\frac{1}{\mathrm{4}}\bigg(p_i^O + p_i^I + p_i^F + p_i^{IF}\bigg), 
    \end{gathered}
  \end{aligned}
\end{equation}
\begin{equation}\label{eq16}
  \begin{aligned}
    \begin{gathered}
      JS\Big(p_{i}^{mean},p_{i}^{perb}\Big)=\frac{1}{2}KLD\left(p_{i}^{mean}||\frac{p_{i}^{mean}+p_{i}^{perb}}{2}\right)\\
      + \frac{1}{2}KLD\left(p_{i}^{perb}||\frac{p_{i}^{perb}+p_{i}^{mean}}{2}\right),perb\in\{O,I,F,IF\},
    \end{gathered}
  \end{aligned}
\end{equation}where $KLD$(.) is the Kullback-Leibler divergence function.

The final loss comprises two components: the classification loss $\mathcal{L}_{cls}$ and the consistency loss $\mathcal{L}_{cons}$, which is defined as:
\begin{equation}\label{eq18}
  \begin{aligned}
    \begin{gathered}
      \mathcal{L}^{perb}_{cls}=\frac{1}{N}\left(\sum_{i=1}^{N}H\left(p_i^{perb},y_i\right)\right),perb\in\{O,I,F,IF\}, \\
      \mathcal{L}^{perb}_{cons}=\frac{1}{N}\Biggl(\sum_{i=1}^NJS\Bigl(p_i^{mean},p_i^{perb}\Bigr)\Biggr),perb\in\{O,I,F,IF\},
    \end{gathered}
  \end{aligned}
\end{equation}where the variable $N$ represents the number of samples in a batch, and $y_i$ represents the label of the $i$-th sample. The standard cross-entropy loss is denoted by $H$.

Then the overall loss function is defined as:
\begin{equation}\label{eq19}
  \begin{aligned}
    \begin{gathered}
        \mathcal{L}=\mathcal{L}_{cls}+\lambda\mathcal{L}_{cons},
    \end{gathered}
  \end{aligned}
\end{equation}where the balance parameter $\lambda$ is fixed at 5 for all datasets, unless specified otherwise. The corresponding pseudo-code is provided in Alg.~\ref{al01}.


\begin{algorithm}[ht]
\caption{Pseudo-code of CPerb in a PyTorch-like style.}
\begin{algorithmic}[1]
\STATE \textcolor{lightgreen}{\# \textbf{f}: The network used for extracting image features.}
\STATE \textcolor{lightgreen}{\# \textbf{aug\_i}: Image-level perturbations.}
\STATE \textcolor{lightgreen}{\# \textbf{f\_p}: Applying MixPatch for feature-level perturbations.}
\vspace{-\baselineskip}
\STATE \textcolor{lightgreen}{\# \textbf{criterion\_c}: Compute the classification loss.}
\STATE \textcolor{lightgreen}{\# \textbf{criterion\_cons}: Compute the loss as Eq.~\ref{eq18}.}
\FOR {x in loader}
\STATE \textcolor{lightgreen}{\# An image-level weak view and an image-level strong view as inputs.}
\STATE $\text{x}^\text{I}$ ~=~ aug\_i(x).
\STATE \textcolor{lightgreen}{\# Utilize the network to generate predicted values. ``f\_p'' represents feature-level perturbation.}
\STATE $\text{p}^{\text{O}}$,~ $\text{p}^{\text{F}}$ ~=~ f(x),~ f(x,~ f\_p=True). 
\STATE $\text{p}^{\text{I}}$,~ $\text{p}^{\text{IF}}$ ~=~ f($\text{x}^\text{I}$),~ f($\text{x}^\text{I}$,~ f\_p=True).
\STATE pred ~=~ [$\text{p}^{\text{O}}$,~ $\text{p}^{\text{I}}$,~ $\text{p}^\text{F}$,~ $\text{p}^{\text{IF}}$].
\STATE \textcolor{lightgreen}{\# Compute prototypes using predicted values.}
\STATE prototype ~=~ torch.stack(pred).mean(dim=0).
\STATE \textcolor{lightgreen}{\# Compute the classification loss.}
\STATE loss\_cl ~=~ criterion\_c(pred,~ label.repeat(4)).
\STATE \textcolor{lightgreen}{\# Measure the similarity between each predicted value and the prototype.}
\FOR {perb in pred}
\STATE loss\_cons ~+=~ criterion\_cons(perb,~ prototype).
\ENDFOR
\STATE \textcolor{lightgreen}{\# Final loss}
\STATE loss ~=~ loss\_cl ~+~ $\lambda$ ~\textasteriskcentered~ loss\_cons.
\ENDFOR
\end{algorithmic}
\label{al01}
\end{algorithm}

\subsection{New Feature Perturbation: MixPatch}\label{SEC:LDII}
\begin{figure}[t]
\centering
\subfigure[Mean and Variance of \underline{means}]{
\begin{minipage}[b]{0.92\linewidth}
\centering
\includegraphics[width=\textwidth]{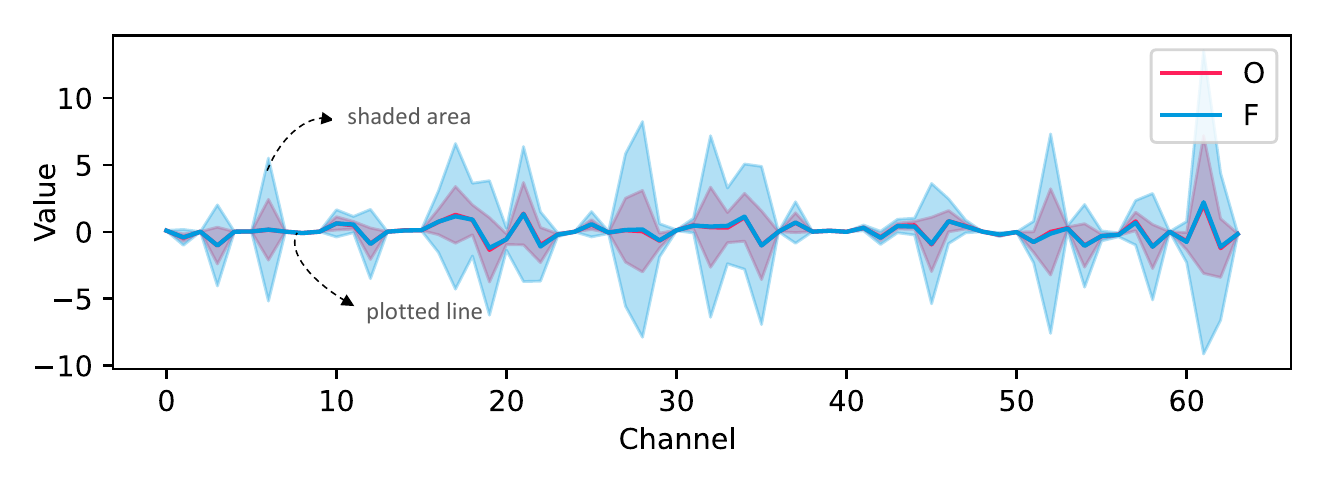}
\end{minipage}
}
\subfigure[Mean and Variance of \underline{variances}]{
\begin{minipage}[b]{0.92\linewidth}
\centering
\includegraphics[width=\textwidth]{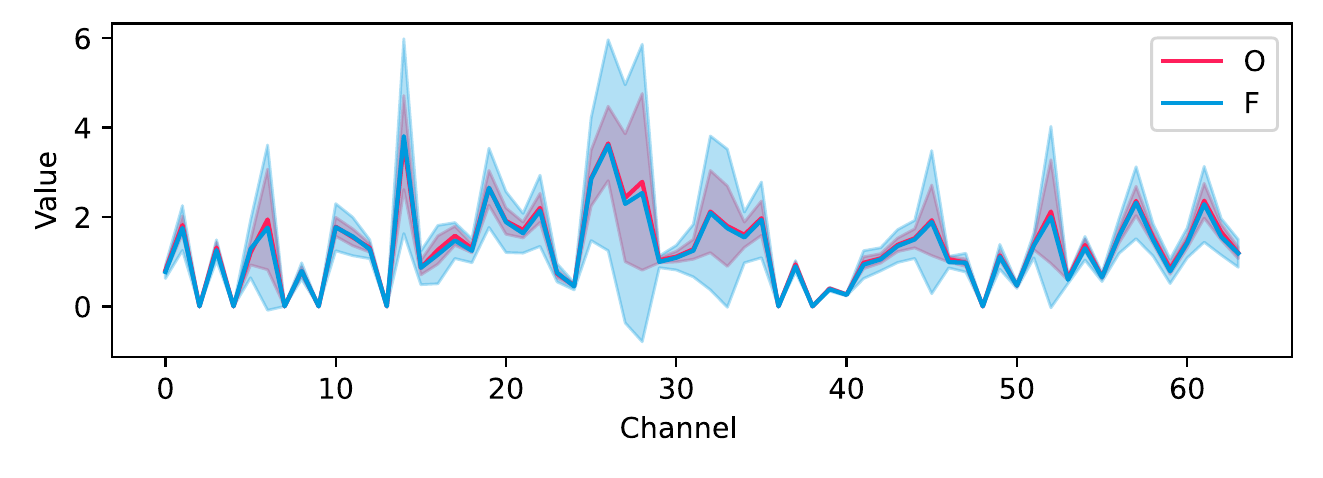}
\end{minipage}
}
\caption{Statistical discrepancies between Original and MixPatch. Experiment on ``art\_painting'' domain of PACS. ``O'' represents the output features obtained after passing the original image through the first convolutional layer of ResNet18. ``F'' corresponds to the output features resulting from applying feature-level perturbation to the original image after feature extraction in the convolutional layer. Mean and variance (\ie, statistics) are computed for features on the original image (O) and feature perturbation (F). Thus, for each channel, ``O'' (``F'') yield 128 (256) means and variances. Then, we compute the mean (\ie, the plotted lines) and variance (\ie, the shaded areas) of 128 (256) means, as shown in (a). Similar to the mean and variance of 128 (256) variances in (b). It is worth noting that ``256'' means we split two patches for each channel of feature maps.}
\label{fig-MixPatch-Motivation}
\vspace*{-15pt}
\end{figure}

Presently, the majority of feature-level perturbation methods focus on individual image channels, neglecting the exploration of local image feature diversity. In Fig.~\ref{fig-MixPatch-Motivation}, our observations reveal that by applying feature perturbation at the patch level within instances, the resulting feature mean and variance demonstrate increased variability, signifying enhanced statistical data scalability. This significant discovery highlights that the MixPatch feature perturbation method encompasses more abundant style information compared to conventional feature perturbation methods. As a consequence, it effectively addresses the limitation of insufficient data diversity during training in the single-source domain generalization.

Our proposed feature-level perturbation method, MixPatch, is based on the statistical difference between non-overlapping patches within an image. To implement MixPatch, we randomly divide the feature map into non-overlapping patches along the $H$ and $W$ dimensions. Let $P$ denote the number of patches, and let $H_p$ and $W_p$ denote the height and width of the $p$-th patch, while $\mu^p$ and $\sigma^p$ represent the mean and standard deviation of the $p$-th patch. The statistical data of the $i$-th channel can then be represented as follows:
\begin{equation}\label{eq20}
  \begin{aligned}
    \begin{gathered}
      \mu_i^p=\frac{1}{H_pW_p}\sum_{h=1}^{H_p}\sum_{w=1}^{W_p}f^p\left[p,i,h,w\right], \\
      \sigma_i^p=\sqrt{\frac{1}{H_pW_p}\sum_{h=1}^{H_p}\sum_{w=1}^{W_p}(f^p[p,i,h,w]-\mu_i^p)^2+\epsilon}\text{.} 
    \end{gathered}
  \end{aligned}
\end{equation}

Hence, the statistical data for the $i$-th channel can be represented as follows:
\begin{equation}\label{eq21}
  \begin{aligned}
    \begin{gathered}
    \mu_{i}:\left(\mu_{i}^{p_{1}}, \mu_{i}^{p_{2}}, \mu_{i}^{p_{3}} \ldots \mu_{i}^{p_{n}}\right), \\\sigma_{i}:\left(\sigma_{i}^{p_{1}}, \sigma_{i}^{p_{2}}, \sigma_{i}^{p_{3}} \ldots \sigma_{i}^{p_{n}}\right),
    \end{gathered}
  \end{aligned}
\end{equation}where $n$ represents the number of patches within channel $i$.

Here, we randomly shuffle the order of patches in  $\mu_{i}$ and $\sigma_{i}$ for the $i$-th channel, and obtain $\bar{\mu}_{i}$ and $\bar{\sigma}_{i}$ as follows:
\begin{equation}\label{eq211}
  \begin{aligned}
    \begin{gathered}
        \bar{\mu}_{i}:\left(\bar{\mu}_{i}^{p_{1}}, \bar{\mu}_{i}^{p_{2}}, \bar{\mu}_{i}^{p_{3}} \ldots \bar{\mu}_{i}^{p_{n}}\right), 
        \\\bar{\sigma}_{i}:\left(\bar{\sigma}_{i}^{p_{1}}, \bar{\sigma}_{i}^{p_{2}}, \bar{\sigma}_{i}^{p_{3}} \ldots \bar{\sigma}_{i}^{p_{n}}\right).
    \end{gathered}
  \end{aligned}
\end{equation}

Following the DSU~\cite{dsu}, we perturb the feature statistics of ${f}^{\bar{p}}$ ($p\in\{p_1, p_2, p_3 \ldots p_n\}$) using Gaussian distribution to generate enriched new feature statistical data information. The variance of ${f}^{\bar{p}}$ feature statistical data can be expressed as:
\begin{equation}\label{eq22}
  \begin{aligned}
    \begin{gathered}
      \Sigma_{\bar{\mu}}^{2}({f}^{\bar{p}}) =\frac{1}{P}\sum_{p=1}^P(\bar{\mu}({f}^{\bar{p}})-\mathbb{E}_p[\bar{\mu}({f}^{\bar{p}})])^2, \\
      \Sigma_{\bar{\sigma}}^2({f}^{\bar{p}}) =\frac{1}{P}\sum_{p=1}^P(\bar{\sigma}({f}^{\bar{p}})-\mathbb{E}_p[\bar{\sigma}({f}^{\bar{p}})])^2,
    \end{gathered}
  \end{aligned}
\end{equation}where $\sum_{\bar{\mu}} \in \mathbb{R}^{C}$ and $\sum_{\bar{\sigma}} \in \mathbb{R}^{C}$ represent the uncertainty estimation of the feature mean ${\bar{\mu}}^{\bar{p}}$ and feature standard deviation ${\bar{\sigma}}^{\bar{p}}$ of the feature ${f}^{\bar{p}}$, respectively.

Once we obtain the variance of the feature statistics data, we proceed to apply Gaussian perturbation to the patch-level feature statistics data of ${f}^{\bar{p}}$. The perturbed feature statistics data of ${f}^{\bar{p}}$ can be represented as follows:
\begin{equation}\label{eq23}
  \begin{aligned}
    \begin{gathered}
      \tilde{\mu}({f}^{\bar{p}}) =\bar{\mu}({f}^{\bar{p}})+\epsilon_{\bar{\mu}}\Sigma_{\bar{\mu}}({f}^{\bar{p}}),\epsilon_{\bar{\mu}}\sim\mathcal{N}(\mathbf{0},\mathbf{1}), \\
      \tilde{\sigma}({f}^{\bar{p}}) =\bar{\sigma}({f}^{\bar{p}})+\epsilon_{\bar{\sigma}}\Sigma_{\bar{\sigma}}({f}^{\bar{p}}),\epsilon_{\bar{\sigma}}\sim\mathcal{N}(\mathbf{0},\mathbf{1}). 
    \end{gathered}
  \end{aligned}
\end{equation}

Lastly, we apply AdaIN~\cite{huang2017arbitrary} to utilize the perturbed feature statistics data and replace the original patch feature statistics data, achieving the desired transformation. The final representation of MixPatch can be expressed as follows:
\begin{equation}\label{eq24}
  \begin{aligned}
  \text{MixPatch}(f^p)=\tilde{\sigma}({f}^{\bar{p}})\frac{f^p-\mu^p}{\sigma^p}+\tilde{\mu}({f}^{\bar{p}}).
  \end{aligned}
\end{equation}

Unlike the DSU, which mainly considers the global feature statistical data of image instances, MixPatch focuses on the local feature statistical data of the feature map. MixPatch aims to enrich the local image features. During training, the feature are randomly divided into two non-equal patches at the instance level. However, during testing, MixPatch is disabled.

\section{Experiments}\label{s-experiment}

In this section, we begin by introducing the experimental datasets, namely CIFAR10-C, CIFAR100-C, and PACS, in Section~\ref{sec:EXP-DS}. We then delve into the experimental details in Section~\ref{sec:EXP-ID}. Moving forward, in Section~\ref{sec:EXP-CWSM}, we perform a comparative analysis between the proposed method and SOTA methods. Then, in order to assess the efficacy of the different components within the proposed framework, we conduct ablation experiments in Section~\ref{sec:EXP-AS}.  Lastly, we further analyze the property of the proposed network and give the visualization results in Section~\ref{sec:EXP-FA}.

\begin{table}
\centering
\caption{Comparison with state-of-the-art (SOTA) methods on PACS. The table presents the results obtained by evaluating our method on different test domains, namely Photo (P), Art (A), Cartoon (C), and Sketch (S). Each result corresponds to the average performance across the specific domain transfer tasks, such as A → P, C → P, and S → P. Bold font and underline are used to indicate the best and second-best results, respectively.}
\setlength{\tabcolsep}{2.6mm}{
\begin{tabular}{l|cccc|c} 
\toprule
\multicolumn{1}{c|}{Methods} & Art           & Cartoon       & Photo         & Sketch        & Avg.            \\ 
\midrule
ERM                         & 68.6          & 70.2          & 40.5          & 36.0          & 53.8           \\
RSC~\cite{huang2020self}                         & 75.3          & 77.4          & 57.2          & 45.9          & 64.0           \\
RSC w/ ASR~\cite{ASR}                     & 76.7          & \underline{79.3}          & 54.6          & 61.6 & 68.0           \\
pAdaIN~\cite{nuriel2021permuted}                      & 60.5          & 68.0          & 35.6          & 30.6          & 48.7           \\
DSU~\cite{dsu}                         & 71.5          & 74.5          & 42.1          & 47.8          & 59.0           \\
DSU w/ MAD~\cite{mad}                  & 72.4          & 74.5          & 44.1          & 49.6          & 60.2           \\
EFDMix~\cite{zhang2022exact}                      & 63.2          & 73.9          & 42.5          & 38.1          & 54.4           \\
XDED~\cite{lee2022cross}                        & 76.5          & 77.2          & 59.1         & 53.1          & 66.5           \\
ADA~\cite{volpi2018generalizing}                         & 59.7          & 67.1          & 33.6          & 27.3          & 46.9           \\
CrossNorm~\cite{tang2021crossnorm}                   & 61.7          & 66.1          & 31.6          & 20.2          & 44.9           \\
ME-ADA~\cite{zhao2020maximum}                      & 59.7          & 67.1          & 33.6          & 26.8          & 46.8           \\
MSAM+RSW~\cite{li2023exploring}                      & 73.1          & \underline{79.3}           & 44.9           & 53.7           & 62.8            \\
CADA~\cite{chen2023center} & 76.3 & 79.1 & 56.6 & 61.6 & \underline{68.4} \\
ALT~\cite{gokhale2023improving}     & 75.7 & 77.3 & 55.1 & 50.7 & 64.7  \\
FACT-DE~\cite{xu2023fourier} & 75.7 & 78.4 & 50.7 & \textbf{62.8} & 66.9 \\

UniMatch~\cite{yang2022revisiting} & \underline{77.9} & 77.8 & 54.8 & 58.2 & 67.2 \\

Pro-RandConv~\cite{choi2023progressive} & 77.0 & 78.5 & \textbf{62.9} & 57.1 & 68.9 \\

ABA~\cite{cheng2023adversarial} & 75.3 & 77.5 & 58.9 & 53.8 & 66.4 \\
\midrule
\rowcolor{gray!20} 
CPerb (ours)          & \textbf{81.7}          & \textbf{81.4} & \underline{62.8} & \underline{62.6} & \textbf{72.1} \\
\bottomrule
\end{tabular}}

\label{tab01}%
\vspace{-10pt}
\end{table}

\subsection{Experimental Settings}\label{sec:EXP-ES}

\subsubsection{Datasets}\label{sec:EXP-DS}
\textbf{CIFAR-10 and CIFAR-100}~\cite{krizhevsky2009learning} are widely used benchmark for evaluating the robustness of classification models. CIFAR-10 and CIFAR-100 consist of small 32 × 32 RGB images, with 50,000 training images and 10,000 testing images. CIFAR-10 includes 10 object categories, while CIFAR-100 has 100 classes. To assess the resilience of models against common corruptions, we utilize CIFAR10-C and CIFAR100-C~\cite{hendrycks2018benchmarking}. These datasets are created by introducing various corruptions to the original CIFAR test sets. Each dataset comprises 19 types of corruptions, such as blur, weather, and digital corruption, with five different severity levels or intensities. The evaluation is conducted by averaging the performance over all corruptions and intensities.
\textbf{PACS}~\cite{li2017deeper} serves as a benchmark for domain generalization tasks. It consists of four domains: art painting, cartoon, photo, and sketch. The dataset comprises a total of 9,991 images distributed across seven object categories. The images are divided as follows: Photo (1,670 images), Art (2,048 images), Cartoon (2,344 images), and Sketch (3,929 images). In our experiments, one domain is chosen as the source domain, while the remaining three domains act as the target domains, resulting in four different tasks with different source domains.

\subsubsection{Implementation Details}\label{sec:EXP-ID}
In the experiments conducted on CIFAR10-C and CIFAR100-C, we adopt the WideResNet~\cite{zagoruyko2016wide} as the backbone network, as previously done by Zhao \textit{et al.}~\cite{zhao2020maximum}. The network is trained using stochastic gradient descent (SGD) with Nesterov momentum optimization, initialized with a learning rate of 0.1. The learning rate is decayed following a cosine annealing schedule~\cite{loshchilov2016sgdr}. To augment the training data, standard random left-right flipping and cropping techniques are applied. The WideResNet model is trained for 100 epochs with a weight decay of 0.0005, following the settings used by Zhao \textit{et al.}~\cite{zhao2020maximum}.


For the PACS experiments, we employ a ResNet-18 network pretrained on ImageNet~\cite{deng2009imagenet} as the backbone, following the setup presented by Huang \textit{et al.}~\cite{huang2020self}, which has demonstrated superior domain generalization performance on PACS. A fully-connected layer is added for classification purposes.

\begin{figure}[t]
\centering
\subfigure[CIFAR10-C]{
\begin{minipage}[b]{0.46\linewidth}
\centering
\includegraphics[width=\textwidth]{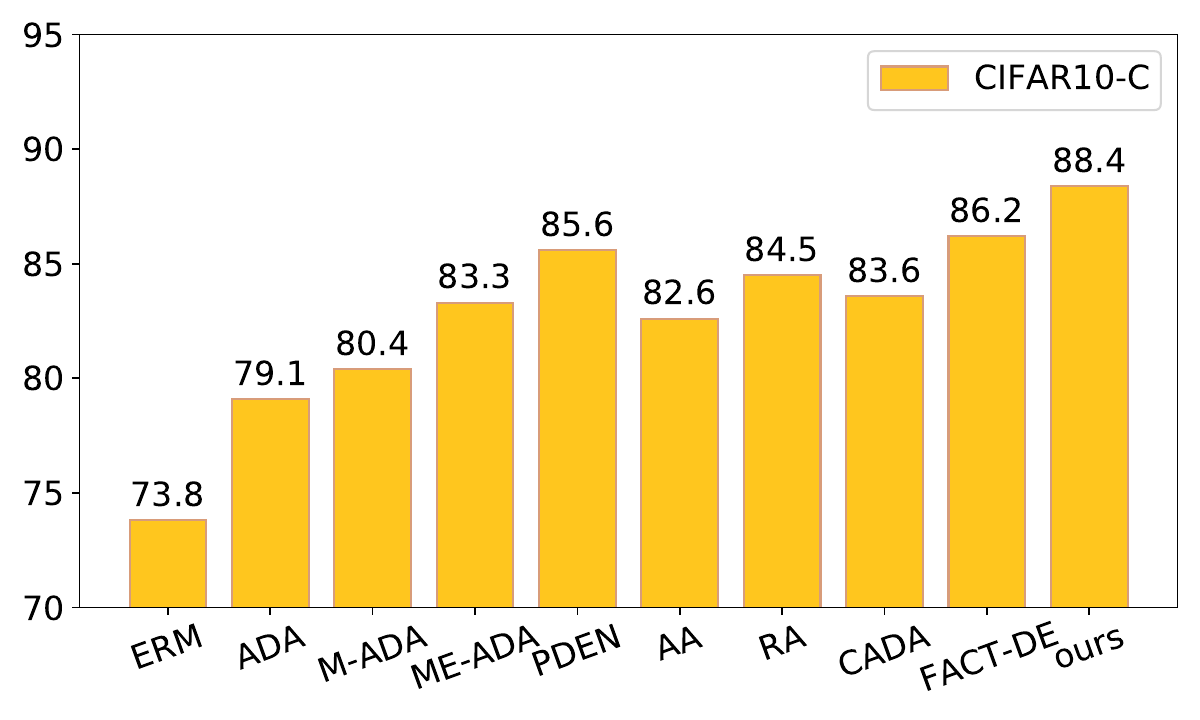}
\end{minipage}
}
\subfigure[CIFAR100-C]{
\begin{minipage}[b]{0.46\linewidth}
\centering
\includegraphics[width=\textwidth]{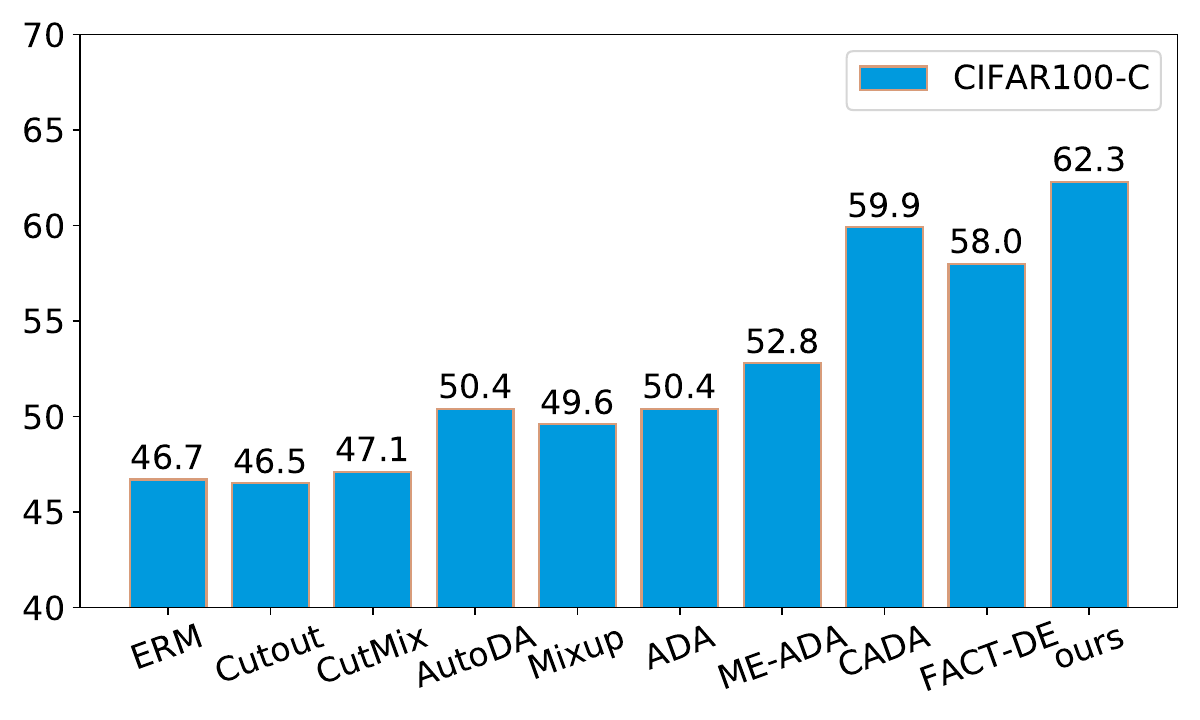}
\end{minipage}
}
\caption{(a) Experimental results of CPerb and SOTA methods in the ``CIFAR-10 → CIFAR10-C'' and ``CIFAR-100 → CIFAR100-C'' tasks.}
\label{fig4}
\vspace*{-15pt}
\end{figure}

\subsection{Comparison with SOTA Methods}\label{sec:EXP-CWSM}
We compare our method with several SOTA methods, including feature-level perturbation methods such as \cite {zhang2022exact, zhou2021domain, li2022uncertainty, zhang2017mixup, devries2017improved, yun2019cutmix}, as well as image-level perturbation methods like \cite{huang2020self, fan2021adversarially, qiao2020learning, volpi2018generalizing, zhao2020maximum, cubuk2019autoaugment, cubuk2020randaugment, li2021progressive}.

Fig.~\ref{fig4} (a), Fig.~\ref{fig4} (b), Tab.~\ref{tab01} illustrate the comparative outcomes of the CPerb and SOTA methods on CIFAR10-C, CIFAR100-C, and PACS, respectively. ``ERM'' stands for the baseline, which utilizes cross-entropy loss on the backbone. Our method is compared with several SOTA methods, including M-ADA~\cite{qiao2020learning}, PDEN~\cite{li2021progressive}, AA~\cite{cubuk2019autoaugment}, RA~\cite{cubuk2020randaugment}, Cutout~\cite{devries2017improved}, CutMix~\cite{yun2019cutmix}, Mixup~\cite{zhang2017mixup}, CADA~\cite{chen2023center}, and FACT-DE~\cite{xu2023fourier}, on the CIFAR10-C and CIFAR100-C datasets.

Based on the aforementioned findings, the following conclusions can be drawn.
Firstly, CPerb consistently outperforms the current SOTA methods across all datasets. In comparison to SOTA methods, our approach demonstrate significant performance gains of $2.2\%$ (CPerb vs. FACT-DE~\cite{xu2023fourier}) and $2.4\%$ (CPerb vs. CADA~\cite{chen2023center}) on CIFAR10-C and CIFAR100-C, respectively. Moreover, we observe a remarkable improvement of $3.7\%$ (CPerb vs. CADA~\cite{chen2023center}) on PACS.
Secondly, when compared to data augmentation techniques closely related to CPerb, CPerb exhibits superior performance over the majority of data augmentation methods, thus reinforcing the superiority of the cross-perturbation framework. 
Furthermore, on PACS characterized by significant domain variations in real-world scenarios, the CPerb framework showcases remarkable performance, achieving an improvement of nearly $18\%$ over the baseline and surpassing the SOTA methods~\cite{huang2020self, fan2021adversarially, chen2023center, lee2022cross, mad, li2023exploring, gokhale2023improving, xu2023fourier} by more than $3\%$. 

\begin{table}[t]
\centering
\caption{Ablation experiments on the image classification PACS dataset. This table presents the results of ablation experiments conducted on the PACS dataset, focusing on different combinations of augmentation techniques. The notation ``I'' denotes image-level perturbation, ``F'' represents feature-level perturbation, and ``IF'' indicates image-feature dual-level perturbation,  where $\mathcal{L}_{cons}$ refers to Eq.~\ref{eq18}.}
\begin{tabular}{cccc|cccc|c} 
\toprule
\multicolumn{4}{c|}{Component} & \multicolumn{4}{c|}{PACS}                                     & \multirow{2}{*}{Avg.}  \\ 
\cline{1-8}
I & IF & F & $\mathcal{L}_{cons}$             & Art           & Cartoon       & Photo         & Sketch        &                       \\ 
\midrule
- & -   & - & -                        & 68.6          & 70.2          & 40.5          & 36.0          & 53.8           \\
- & -  & \checkmark  & -                & 76.1          & 76.9          & 52.5          & 43.6          & 62.3                  \\
\checkmark & -  & -  & -                & 74.7          & 76.2          & 46.6          & 50.2          & 61.9                  \\
- & \checkmark  & -  & -                & 78.8          & 80.9          & 59.3          & 57.4          & 69.1                  \\
\checkmark & -  & \checkmark  & -                & 79.2          & 80.5          & 58.9          & 56.8          & 68.9                  \\
\checkmark & -  & \checkmark  & \checkmark                & 80.0          & 81.4          & \underline{61.9}  & 59.8          & 70.8                  \\
\checkmark & \checkmark  & -  & -                & 80.3          & 80.7          & 59.7          & 59.5          & 70.0                  \\
\checkmark & \checkmark  & -  & \checkmark                & \underline{81.3}  & 80.8          & 61.3          & \underline{60.5}  & 71.0                  \\
- & \checkmark  & \checkmark  & -                & 79.9          & 81.1          & 60.7          & 58.6          & 70.1                  \\
- & \checkmark  & \checkmark  & \checkmark                & 80.1          & \textbf{81.5} & \textbf{62.8} & 60.1          & \underline{71.2}          \\
\checkmark & \checkmark  & \checkmark  & -                & 79.8          & 81.2          & 60.8          & 59.9          & 70.4                  \\
\rowcolor{gray!20} 
\checkmark & \checkmark  & \checkmark  & \checkmark                & \textbf{81.7} & \underline{81.4}  & \textbf{62.8} & \textbf{62.6} & \textbf{72.1}         \\
\bottomrule
\end{tabular}
\label{tab-pacs}%
\vspace{-15pt}
\end{table}

These results further validate the efficacy of the CPerb cross-perturbation framework in addressing domain variations in real-world settings, enhancing the capacity of the model to learn domain-invariant features, and mitigating the risk of overfitting to the source domain. 
Moreover, in experiments conducted on real-world PACS, the most substantial disparity is observed between the Photo domain and the other three domains, leading to a significant decrease in generalization performance when solely trained on the Photo domain. However, the CPerb demonstrates an average precision improvement of approximately $3.7\%$ compared to the XDED~\cite{lee2022cross} when trained on the Photo domain, further underscoring its superior performance in scenarios with broader domain spans. This enhancement can be attributed to the multi-route promotion of domain-invariant feature learning and the complementary nature of image-level and feature-level perturbations, effectively addressing the issue of insufficient training data in single-source domain generalization.


\subsection{Ablation Studies} \label{sec:EXP-AS}
In order to assess the efficacy of each module within our method, we conducte thorough ablation experiments on several single domain generalization tasks. These experiments are designed to explore two key aspects: horizontal and vertical directions. Firstly, in the vertical direction, we present empirical evidence demonstrating the benefits of employing multi-route learning techniques at both the image and feature levels, surpassing the performance of single-route learning methods. Secondly, in the horizontal direction, we introduce the image-feature dual-level complementary module to demonstrate its superior performance, while also highlighting the distinctions between the MixPatch feature-level perturbation method and previous methods used for feature perturbation.

\begin{figure}[t]
\centering
\subfigure[MulMs]{
\begin{minipage}[b]{0.46\linewidth}
\centering
\includegraphics[width=\textwidth]{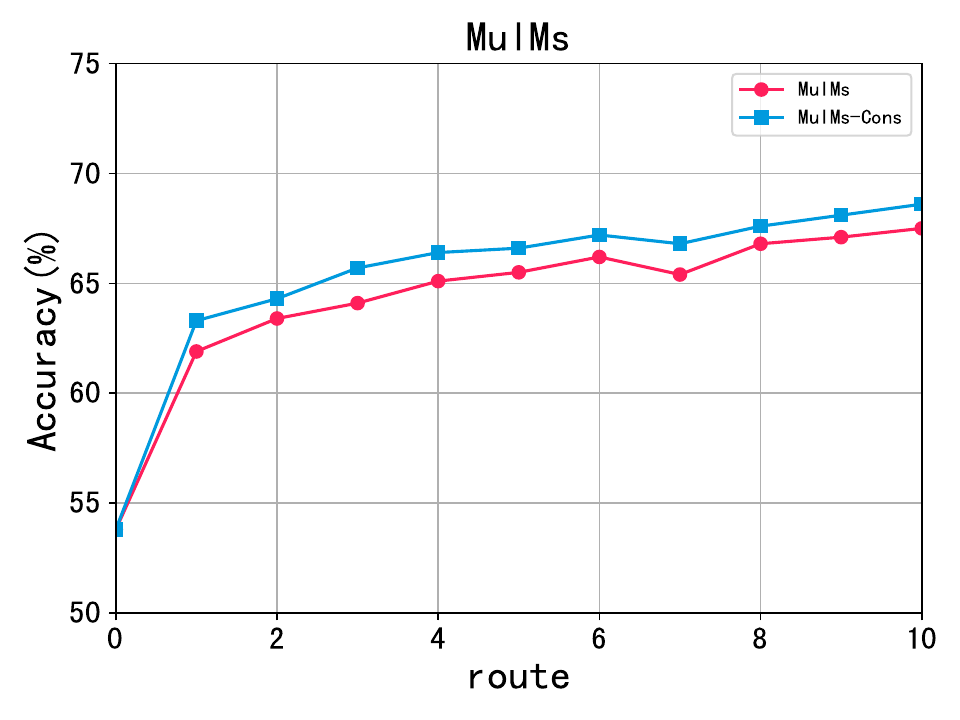}
\end{minipage}
}
\subfigure[MulFp]{
\begin{minipage}[b]{0.46\linewidth}
\centering
\includegraphics[width=\textwidth]{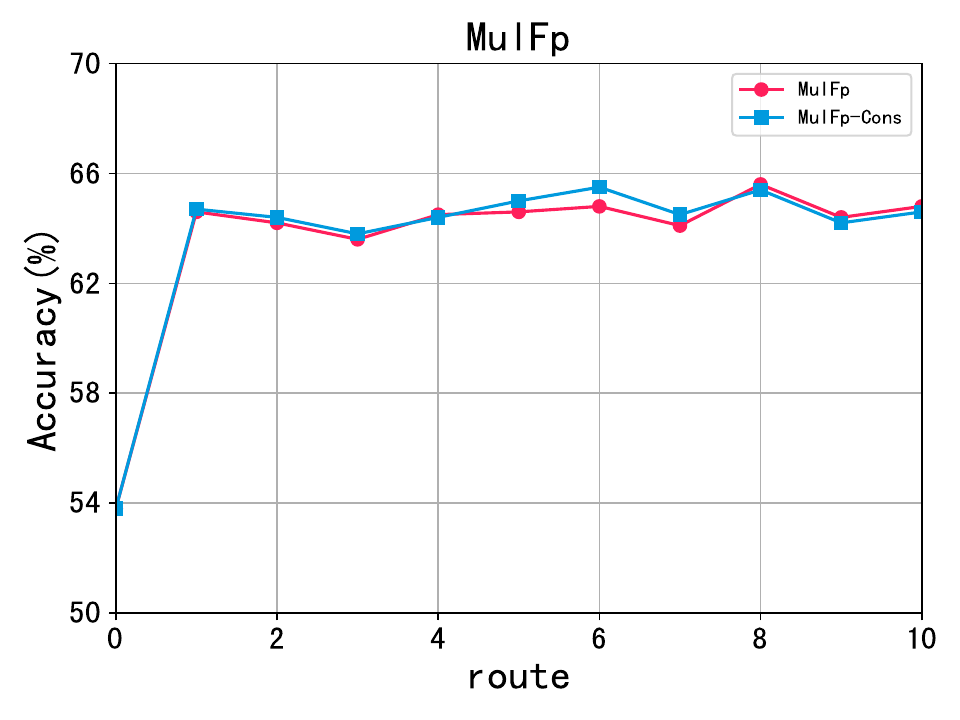}
\end{minipage}
}
\caption{(a) Ablation experiments for MulMs framework on PACS. (b) Ablation experiments for MulFp framework on PACS. The horizontal axis, labeled as ``route'', signifies the count of perturbations. In this context, the blue line illustrates the scenario in which consistency loss is not employed, while the red line portrays the scenario in which consistency loss is incorporated.}
\label{fig6}
\vspace*{-20pt}
\end{figure}


\subsubsection{\textbf{Analysis of multi-route feature perturbation (MulFp)}} \label{sec:EXP-ASMF}
In this experiment, we explore the influence of multi-route feature perturbation in the vertical direction, employing our proposed MixPatch as the feature perturbation method. The experimental results in Fig.~\ref{fig6} (a) yield the following insights. Firstly, by introducing an additional route of feature-level perturbation alongside original image, we observe an $8.8\%$ improvement over the baseline. This finding suggests that MixPatch, the feature-level perturbation method, effectively complements original image, thereby enhancing the diversity of training samples and addressing the issue of insufficient training data in single-source domain generalization. 

Secondly, when employing multi-route feature-level perturbation, the performance exhibits minor fluctuations while remaining relatively stable. This behavior can be attributed to the fact that MixPatch randomly swaps patch styles within each batch, effectively enriching the diversity of sample features and achieving the desired effects of multi-route perturbation. Consequently, the introduction of additional feature perturbation routes does not significantly augment sample diversity beyond a certain point. Moreover, incorporating consistency loss on top of multi-route feature-level perturbation yields only marginal improvement. This observation arises from the fact that single-route feature perturbation already enhances feature diversity internally, and the distinctions between multiple routes of feature perturbation are not substantial. Consequently, ensuring consistency among multiple routes does not yield a significant improvement.



\begin{table*}[t]
\centering
\caption{Ablation Experiments on the CIFAR10-C and CIFAR100-C Image Classification Dataset.}
\begin{tabular}{cccc|ccccc|>{\columncolor[gray]{1}}c|ccccc|>{\columncolor[gray]{1}}c} 
\toprule
\multicolumn{4}{c|}{Component} & \multicolumn{6}{c|}{CIFAR10-C}                                                                & \multicolumn{6}{c}{CIFAR100-C}                                                                                                                                     \\ 
\midrule
I & IF & F & $\mathcal{L}_{cons}$             & Level 1       & Level 2       & Level 3       & Level 4       & \multicolumn{1}{c|}{Level 5}       & Avg.          & Level 1       & \multicolumn{1}{c}{Level 2} & \multicolumn{1}{c}{Level 3} & \multicolumn{1}{c}{Level 4} & \multicolumn{1}{c|}{Level 5} & Avg.  \\ 
\midrule
  - & -   & -  & -                 & 87.8           & 81.5           & 75.5           & 68.2           & 56.1           & 73.8           & 61.3           & 53.2                         & 47.1                         & 41.2                         & 30.9                          & 46.7                       \\
  - & -   & \checkmark  & -                 & 89.2          & 84.1          & 78.6          & 71.5          & 59.9          & 76.7          & 65.1          & 56.9                        & 50.9                        & 44.5                        & 34.2                         & 50.3                      \\
\checkmark &-    &-    &-                  & 91.7          & 89.3          & 86.5          & 83.0          & 77.8          & 85.7          & 68.1          & 65.1                        & 61.3                        & 56.4                        & 49.1                         & 60.0                      \\
  -& \checkmark  &-    &-                  & 91.7          & 89.6          & 87.2          & 83.8          & 78.4          & 86.1          & 68.7          & 65.3                        & 61.6                        & 56.9                        & 49.8                         & 60.5                      \\
\checkmark &-    & \checkmark  &-                  & \underline{92.3}  & 90.0          & 87.6          & 83.6          & 78.3          & 86.4          & \underline{70.0}  & 65.3                        & 61.6                        & 56.6                        & 49.3                         & 60.6                      \\
\checkmark & -   & \checkmark  & \checkmark                & \textbf{92.4} & \underline{90.6}  & \underline{88.4}  & 85.2          & 80.3          & 87.4          & \textbf{70.1} & \textbf{66.5}               & \underline{63.1}                & \underline{58.8}                & 52.3                         & \underline{62.2}              \\
\checkmark & \checkmark  & -   &   -               & 91.9          & 90.0          & 88.0          & 84.9          & 80.5          & 87.1          & 68.5          & 64.8                        & 61.6                        & 56.8                        & 50.1                         & 60.4                      \\
\checkmark & \checkmark  & -   & \checkmark                & 91.8          & 90.2          & 88.3          & \underline{85.7}  & 81.7          & 87.5          & 68.6          & 65.4                        & 62.5                        & 58.4                        & \underline{52.5}                 & 61.5                      \\
  -& \checkmark  & \checkmark  &  -                & 91.7          & 89.3          & 87.4          & 84.1          & 79.2          & 86.3          & 69.1          & 64.4                        & 60.6                        & 55.5                        & 47.4                         & 59.4                      \\
  -& \checkmark  & \checkmark  & \checkmark                & 91.8          & 89.9          & 87.8          & 84.6          & 79.6          & 86.7          & 69.3          & 65.5                        & 62.1                        & 57.3                        & 49.7                         & 60.8                      \\
\checkmark & \checkmark  & \checkmark  & -                 & 92.1          & 90.1          & \underline{88.4}  & 85.3          & \underline{82.4}  & \underline{87.7}  & 69.4          & 65.4                        & 61.7                        & 57.9                        & 52.2                         & 61.3                      \\
\rowcolor{gray!20} 
\checkmark & \checkmark  & \checkmark  & \checkmark                & \underline{92.3}  & \textbf{90.6} & \textbf{89.0} & \textbf{86.8} & \textbf{83.1} & \textbf{88.4} & 69.9          & \underline{66.0}                & \textbf{63.3}               & \textbf{59.1}               & \textbf{53.3}                & \textbf{62.3}             \\
\bottomrule
\end{tabular}
\label{tab-cifar}%
\vspace{-10pt}
\end{table*}

\subsubsection{\textbf{Analysis of multi-route image perturbation (MulMs)}} \label{sec:EXP-ASMS}
In this experiment, we continue our investigation into the benefits of vertical multi-route image perturbation. The experimental results in Fig.~\ref{fig6} (b) reveal several key findings. Firstly, including image-level augmentation leads to a notable $7.9\%$ improvement compared to its absence. This demonstrates the significant enhancement in sample diversity achieved by image-level augmentation, effectively addressing the challenge of limited training data in single-source domain generalization with significant domain shifts. Secondly, the performance improves as we employ more routes of image-level augmentation. This improvement is attributed to the increased diversity introduced by the stochastic nature of image-level augmentation, further enriching sample diversity. Additionally, the introduction of consistency discrepancy among multiple routes significantly outperforms the MulFp. This observation arises from the substantial difference among multiple routes of image-level perturbation. Ensuring consistency in learning features for different views of the same image promotes the extraction of domain-invariant features.

\begin{table}
\centering
\caption{Comparative Analysis of Various Feature Perturbation Methods.}

\begin{tabular}{l|cccc|c} 
\toprule
\multicolumn{1}{c|}{Methods} & Art           & Cartoon       & Photo         & Sketch        & Avg.            \\ 
\midrule
ERM                         & 68.6          & 70.2          & 40.5          & 36.0          & 53.8           \\
CPerb (Mixstyle~\cite{zhou2021domain})           & 81.5          & 80.1          & 58.9          & 58.2          & 69.7           \\
CPerb (EFDMix~\cite{zhang2022exact})             & \textbf{82.1} & 80.5          & 61.3          & 55.9          & 70.0           \\
CPerb (DSU~\cite{dsu})                & 81.2          & 80.5       & 60.3          & 59.0 & 70.3           \\

CPerb (CPSS~\cite{zhao2022source}) & 78.8 & \underline{80.8} & \underline{61.9} & \underline{61.4} & \underline{70.7} \\
\rowcolor{gray!20} 
CPerb (MixPatch)          & \underline{81.7}          & \textbf{81.4} & \textbf{62.8} & \textbf{62.6} & \textbf{72.1}  \\
\midrule

CPerb(Mixstyle+MixPatch) & 81.4 & 80.9 & 62.0 & 62.8 & 71.8 \\

CPerb(EFDMix+MixPatch) & \textbf{81.9} & 80.7 & 62.3 & 62.5 & 71.9 \\

CPerb(DSU+MixPatch) & 81.5 & \textbf{81.4} & \textbf{63.4} & \textbf{63.3} & \textbf{72.4} \\
\bottomrule
\end{tabular}
\label{tab03}%
\vspace{-15pt}
\end{table}

\subsubsection{\textbf{Ablation studies for CPerb}} \label{sec:EXP-ASCP}
In this experiment, we delve into the advantages of the horizontal image-feature dual-level method and vertical multi-route perturbation in this experiment. The CPerb framework encompasses four routes: original image (O), image-level perturbation (I), MixPatch feature-level perturbation (F), and image-feature dual-level perturbation (IF). 
Notable observations can be made through the ablation experiments conducted on the PACS dataset, as presented in Tab.~\ref{tab-pacs}. Individually adding I, F, and IF on top of the baseline results in performance improvements of $8.1\%$, $8.5\%$, and $15.3\%$, respectively. This confirms the effectiveness of image-feature dual-level perturbation in enhancing data diversity at the image and feature levels. Simultaneously incorporating I, F, and IF on top of the baseline leads to a remarkable $16.6\%$ improvement. This enhancement is attributed to the beneficial effect of employing multi-route perturbations, which aids in the acquisition of domain-invariant features. Additionally, the introduction of consistency loss for learning domain-invariant features further improves performance by $1.3\%$. This arises from the significant differences introduced among multiple routes through the combination of feature-level perturbation and image-level perturbation, with the JS discrepancy reducing sensitivity to style information and facilitating the learning of domain-invariant features. Moreover, additional ablation experiments are carried out on the CIFAR10-C and CIFAR100-C datasets, as illustrated in Tab.~\ref{tab-cifar}, providing supplementary evidence to reinforce our previous findings.


\begin{figure}[t]
\centering
\subfigure[Mean and Variance of \underline{means}]{
\begin{minipage}[b]{0.8\linewidth}
\centering
\includegraphics[width=\textwidth]{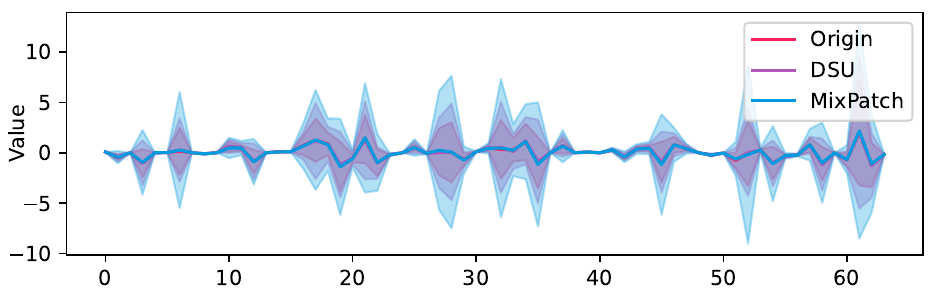}
\end{minipage}
}
\subfigure[Mean and Variance of \underline{variances}]{
\begin{minipage}[b]{0.8\linewidth}
\centering
\includegraphics[width=\textwidth]{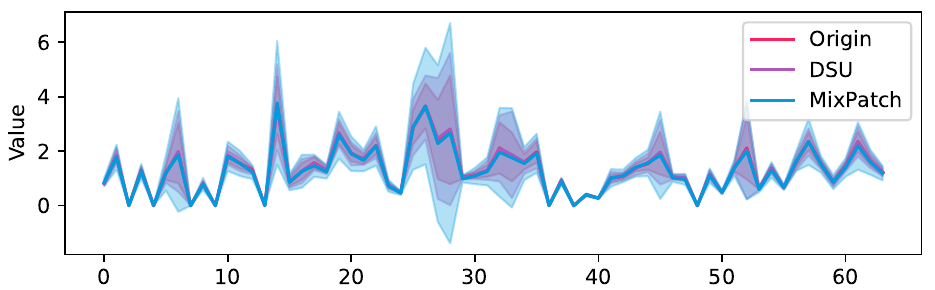}
\end{minipage}
}
\caption{Statistical discrepancies between DSU~\cite{dsu} and MixPatch. Experiment on ``art\_painting'' domain of PACS. ``Origin'' represents the output features obtained after passing the original image through the first convolutional layer of ResNet18. ``MixPatch'' corresponds to the output features resulting from applying MixPatch perturbation to the original image after feature extraction in the convolutional layer. ``DSU'' corresponds to the output features resulting from applying DSU perturbation to the original image after feature extraction in the convolutional layer. Mean and variance (\ie, statistics) are computed for features on the MixPatch and DSU. Thus, for each channel, ``DSU'' (``MixPatch'') yield 128 (256) means and variances. Then, we compute the mean (\ie, the plotted lines) and variance (\ie, the shaded areas) of 128 (256) means, as shown in (a). Similar to the mean and variance of 128 (256) variances in (b). It is worth noting that ``256'' means we split two patches for each channel of feature maps.}
\label{fig-MixPatch-DSU}
\vspace*{-24pt}
\end{figure}

\subsection{Further Analysis} \label{sec:EXP-FA}
\subsubsection{\textbf{Effect of MixPatch}} \label{sec:EXP-FAMP}
Moreover, we thoroughly examine the differences between various feature perturbation methods in this experiment. 
The MixPatch feature perturbation method is compared with current SOTA methods within the CPerb framework in Table~\ref{tab03}. The experimental results consistently demonstrate the superiority of the MixPatch feature perturbation method. Existing feature perturbation methods primarily focus on style swapping between image instances within the same batch or random perturbation based on the original instances, overlooking the style differences among different local patches within an instance. In contrast, our MixPatch further enhances training data diversity by performing style swapping among local patches.\\
As presented in Tab.~\ref{tab03}, experiments are conducted by integrating various feature perturbation methods. It is evident that combining the MixPatch method with other perturbation methods has minimal impact on the final results, consistent with the observations from Fig.~\ref{fig-MixPatch-DSU}. Fig.~\ref{fig-MixPatch-DSU} illustrates that the feature perturbation range of MixPatch essentially encompasses that of DSU, underscoring the rich style generation capabilities of the MixPatch feature perturbation method and affirming its effectiveness.

\subsubsection{\textbf{Different splitting scheme in MixPatch}} \label{sec:EXP-FADSS}
Additionally, we present the results of different partitioning methods in Tab.~\ref{tab04}. To avoid excessively small patches, we perform partitioning at 1/3 and 2/3 of the width or height when dividing into two and four blocks, respectively. For nine blocks, we choose two positions (from 1/5 to 2/5 and from 3/5 to 4/5) for partitioning. The experimental results indicate that random vertical partitioning yields the best performance. It is important to note that unequal partitioning into nine blocks leads to training instability due to the random allocation of patches of varying sizes, which can result in some patches being too small and causing numerical instability.

\begin{figure}[t]
\centering
\subfigure[]{
\begin{minipage}[b]{0.46\linewidth}
\centering
\includegraphics[width=\textwidth]{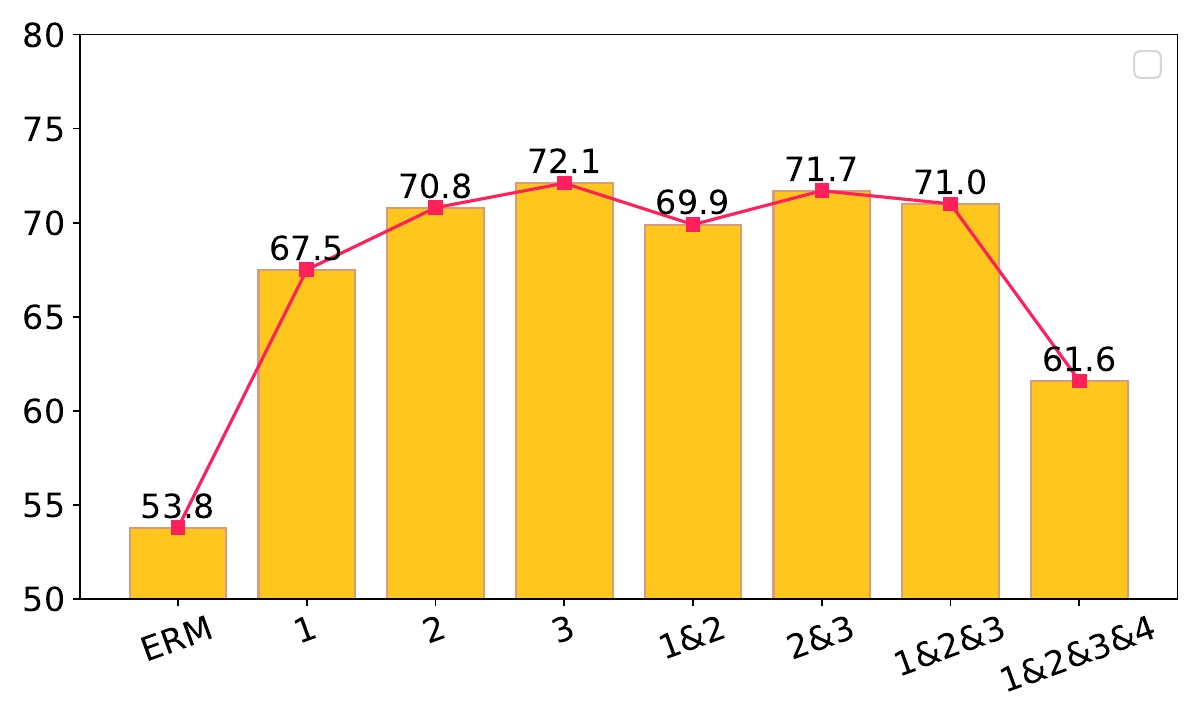}
\end{minipage}
}
\subfigure[]{
\begin{minipage}[b]{0.46\linewidth}
\centering
\includegraphics[width=\textwidth]{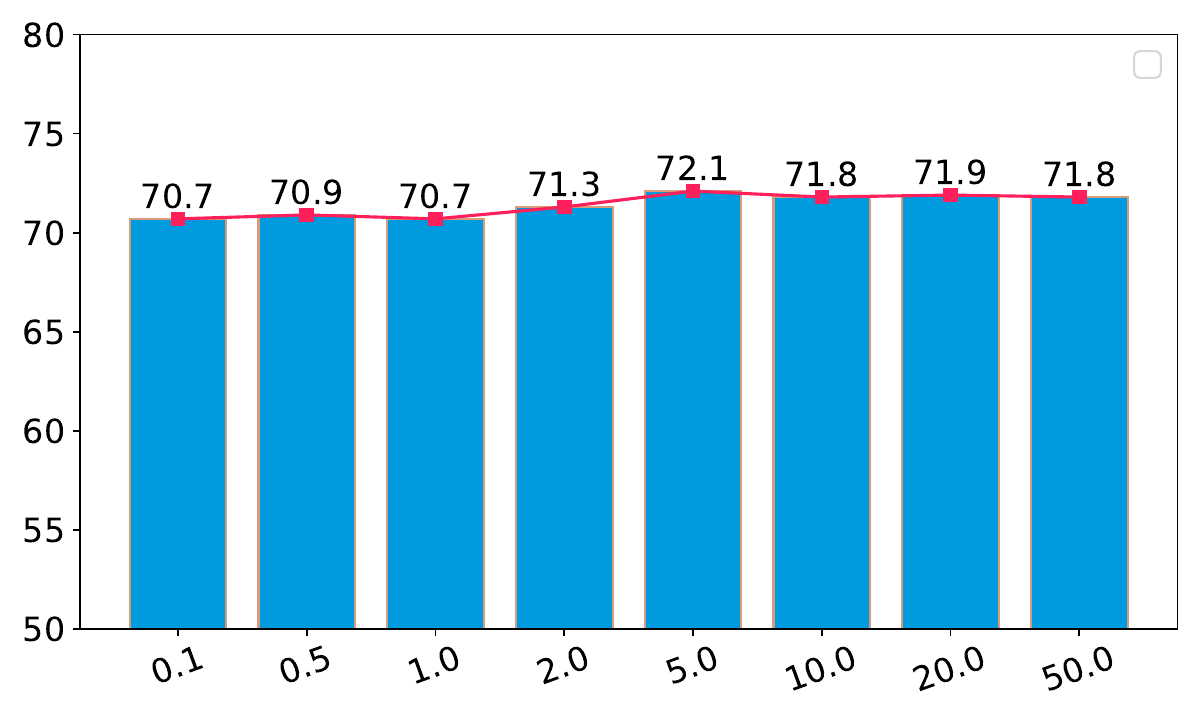}
\end{minipage}
}
\caption{(a) Application of MixPatch feature perturbation method at different positions in the ResNet-18 backbone network. (b) Analysis of the hyperparameter $\lambda$ in Eq.~\ref{eq19}.}
\label{fig-layer-lambda}
\vspace*{-18pt}
\end{figure}



\subsubsection{\textbf{Analysis of the hyper-parameter}} \label{sec:EXP-FAHP}
In Eq.~\ref{eq19}, the hyperparameter $\lambda$ is utilized to strike a balance between the classification loss and the consistency loss. We investigate the influence of different values of $\lambda$, and the experimental outcomes are showcased in Fig.~\ref{fig-layer-lambda} (b). Based on the findings obtained from the PACS dataset, several noteworthy observations can be made. Firstly, the optimal performance is attained when $\lambda$ is set to $5$, resulting in an accuracy of $72.1\%$. Secondly, across the $\lambda$ range of $0.1$ to $50$, there is no substantial disparity in experimental accuracy. This suggests that our CPerb exhibits robustness to variations in the $\lambda$ parameter.



\begin{table}
\centering
\caption{Results of Various Splitting Schemes and Patch Numbers on PACS. This table presents the experimental results of different splitting schemes and the number of patches applied to the PACS dataset. The notation ``P2'' signifies the splitting of feature maps into two patches. ``LR'' and ``UD'' denote the splitting of feature maps into left-right and up-down patches, respectively. ``Equal'' indicates that all patches are of equal size, while ``random'' signifies the random division of feature maps into different patches. ``N/A'' denotes non-convergence of the model during training.}

\setlength{\tabcolsep}{2.8mm}{
\begin{tabular}{l|cccc|c} 
\toprule
\multicolumn{1}{c|}{Methods} & Art           & Cartoon       & Photo         & Sketch        & Avg.            \\ 
\midrule
P2-LR-equal                & \underline{82.2}          & \textbf{81.5} & 62.5          & 61.1          & \underline{71.8}           \\
P2-UD-equal                & 81.5          & 81.0          & 62.1          & \underline{61.4}          & 71.5           \\
P2-LR-random               & \textbf{83.1} & \textbf{81.5} & \textbf{63.2} & 58.3          & 71.5           \\
\rowcolor{gray!20} 
P2-UD-random               & 81.7          & \underline{81.4}          & \underline{62.8}          & \textbf{62.6} & \textbf{72.1}  \\
P4-equal                   & 78.8          & 80.8          & 61.9          & \underline{61.4}          & 70.7           \\
P4-random                  & 67.2          & 80.4          & 60.4          & 51.9          & 65.0           \\
P9-equal                   & 52.8          & 53.0          & 32.3          & 34.1          & 43.1           \\
P9-random                  & N/A           & N/A           & N/A           & N/A           & N/A \\
\bottomrule
\end{tabular}}
\label{tab04}%
\vspace{-10pt}
\end{table}

\begin{table}
\centering
\caption{Experimental Results on the PACS Dataset using ResNet-50 as the Backbone Network.}

\setlength{\tabcolsep}{2.4mm}{
\begin{tabular}{l|cccc|c} 
\toprule
\multicolumn{1}{c|}{Methods} & Art           & Cartoon       & Photo         & Sketch        & Avg.            \\ 
\midrule
ERM                         & 71.8          & 76.1          & 44.6          & 36.0          & 57.1           \\
MixStyle~\cite{zhou2021domain}      & 73.2          & 74.8          & 46.0          & 40.6          & 58.6           \\
EFDMix~\cite{zhang2022exact}                      & {75.3}          & {77.4}          & {48.0}          & {44.2}          & {61.2}           \\
ASA~\cite{zhang2023adversarial}	&  \underline{77.3} 	 & \underline{78.8} 	 & \underline{50.3} 	 & \underline{61.8} 	& 67.0 \\
Pro-RandConv~\cite{choi2023progressive}	& -	& -	& -	& -	&  \underline{73.3} \\
\midrule
\rowcolor{gray!20} 
CPerb (ours)                        & \textbf{85.1} & \textbf{86.1} & \textbf{66.6} & \textbf{72.6} & \textbf{77.6}  \\
\bottomrule
\end{tabular}
}
\label{tab07}%
\vspace{-15pt}
\end{table}

\begin{table*}[th]
\centering
\caption{Comparison with SOTA methods in the person re-identification task. All experimental results are averaged over five repeated trials.}
\setlength{\tabcolsep}{4mm}{
\begin{tabular}{l|cccc|cccc} 
\toprule
\multirow{2}[1]{*}{Methods} & \multicolumn{4}{c|}{Market1501$\to$GRID}                                                       & \multicolumn{4}{c}{GRID$\to$Market1501}                                                       \\ 
\cmidrule{2-9}    \multicolumn{1}{c|}{}
           & mAP               & R1                & R5                & R10               & mAP              & R1                & R5                & R10                \\ 
\midrule
OSNet                                       & 33.3$\pm$0.4          & 24.5$\pm$0.4          & 42.1$\pm$1.0          & 48.8$\pm$0.7          & 3.9$\pm$0.4          & 13.1$\pm$1.0          & 25.3$\pm$2.2          & 31.7$\pm$2.0           \\
MixStyle~\cite{zhou2021domain}                                    & 33.8$\pm$0.9          & 24.8$\pm$1.6          & 43.7$\pm$2.0          & 53.1$\pm$1.6          & 4.9$\pm$0.2          & 15.4$\pm$1.2          & 28.4$\pm$1.3          & 35.7$\pm$0.9           \\
EFDMix~\cite{zhang2022exact}                                      & \underline{35.5$\pm$1.8}          & \underline{26.7$\pm$3.3}          & \underline{44.4$\pm$0.8}          & \underline{53.6$\pm$2.0}          & \underline{6.4$\pm$0.2}          & \underline{19.9$\pm$0.6}          & \underline{34.4$\pm$1.0}          & \underline{42.2$\pm$0.8}           \\
\midrule
\rowcolor{gray!20} 
CPerb (ours)                                     & \textbf{39.3$\pm$1.8} & \textbf{29.3$\pm$1.0} & \textbf{49.6$\pm$3.0} & \textbf{59.2$\pm$4.7} & \textbf{7.2$\pm$0.2} & \textbf{22.2$\pm$0.5} & \textbf{38.4$\pm$0.9} & \textbf{46.4$\pm$0.8}  \\
\bottomrule
\end{tabular}}
\label{tab08}%
\vspace{-20pt}
\end{table*}

\subsubsection{\textbf{Evaluation of the impact of our method at different positions}} \label{sec:EXP-FADP}
To assess the effectiveness of the MixPatch at various positions within the network, we present the experimental results obtained by applying the MixPatch module in different layers, as illustrated in Fig.~\ref{fig-layer-lambda} (a). Notably, two key observations can be made from this figure. Firstly, applying the MixPatch feature perturbation method in the third layer yields the most favorable outcomes. Secondly, employing MixPatch feature perturbation in the shallower layers generally leads to superior performance. This discrepancy arises from the fact that the shallow layers of deep networks encompass more image style information, whereas the deeper layers contain a greater amount of semantic information.

\subsubsection{\textbf{Results based on ResNet-50}} \label{sec:EXP-FAR50}
To validate the effectiveness of the CPerb framework, we perform additional comparative experiments utilizing ResNet-50 as the backbone on the PACS dataset. The experimental results are presented in Tab.~\ref{tab07}. Notably, CPerb exhibits substantial improvements over the baseline when employed with ResNet-50, further highlighting the superiority of the CPerb framework.






\begin{figure}[t]
\centering
\begin{minipage}[t]{1.0\linewidth}
\centering
\subfigure[Photo]{
  \includegraphics[width=0.21\linewidth]{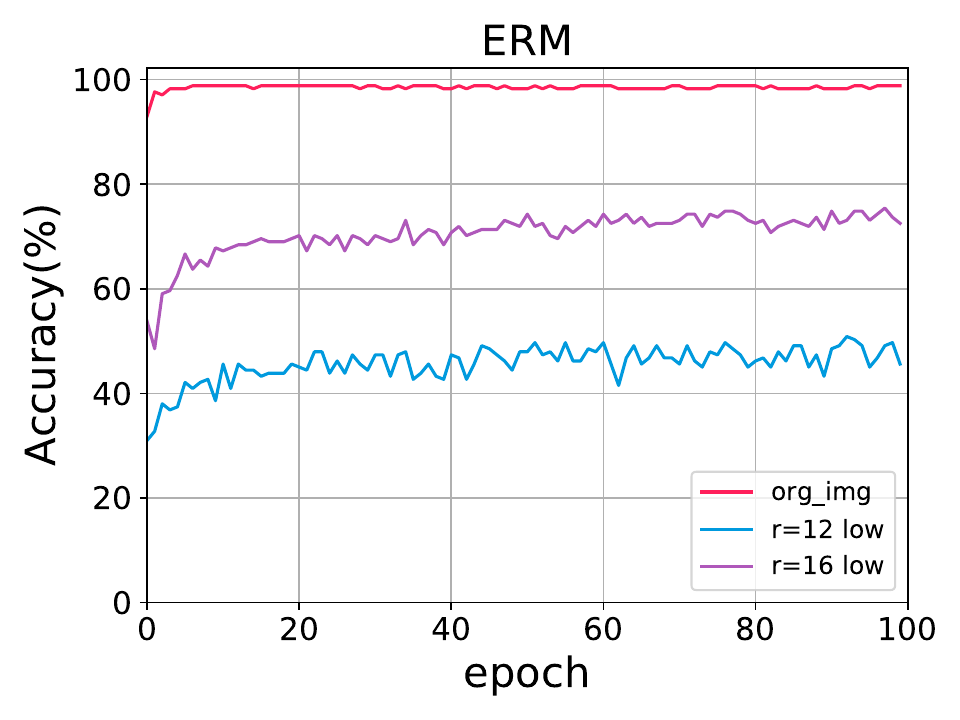}
}
\subfigure[Cartoon]{
  \includegraphics[width=0.21\linewidth]{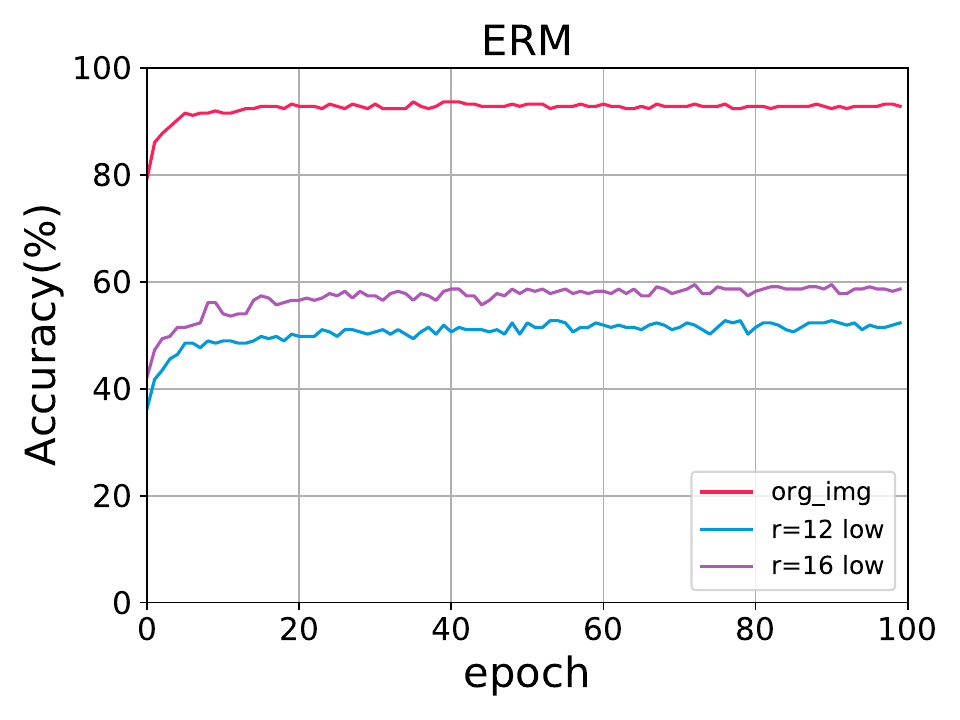}
}
\subfigure[Art]{
  \includegraphics[width=0.21\linewidth]{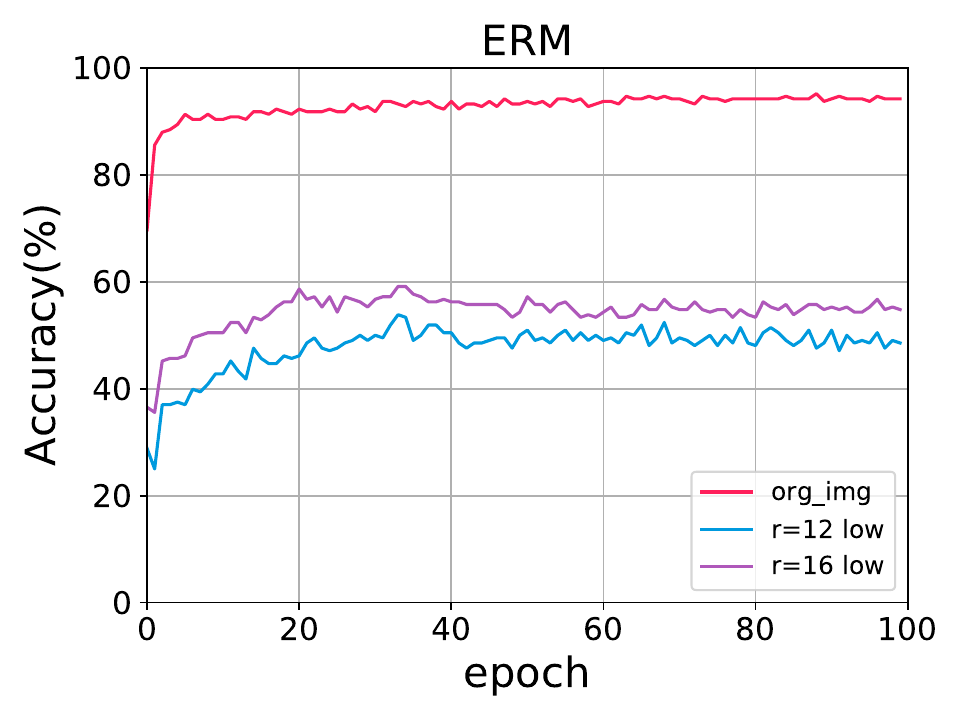}
}
\subfigure[Sketch]{
  \includegraphics[width=0.21\linewidth]{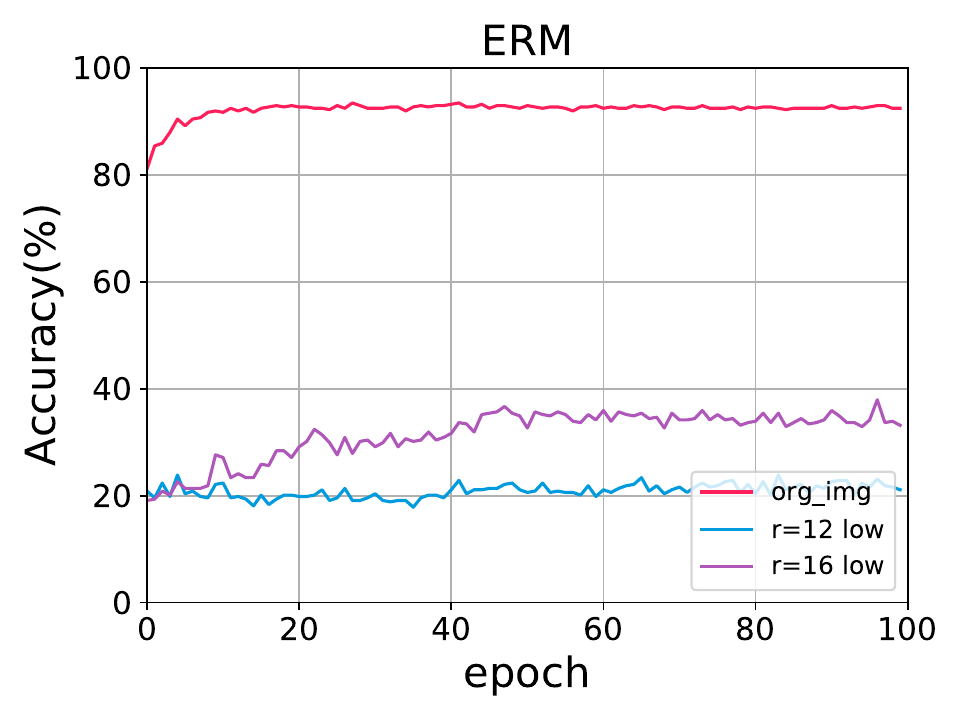}
}

\end{minipage}%

\begin{minipage}[t]{1.0\linewidth}
\centering
\subfigure[Photo]{
  \includegraphics[width=0.21\linewidth]{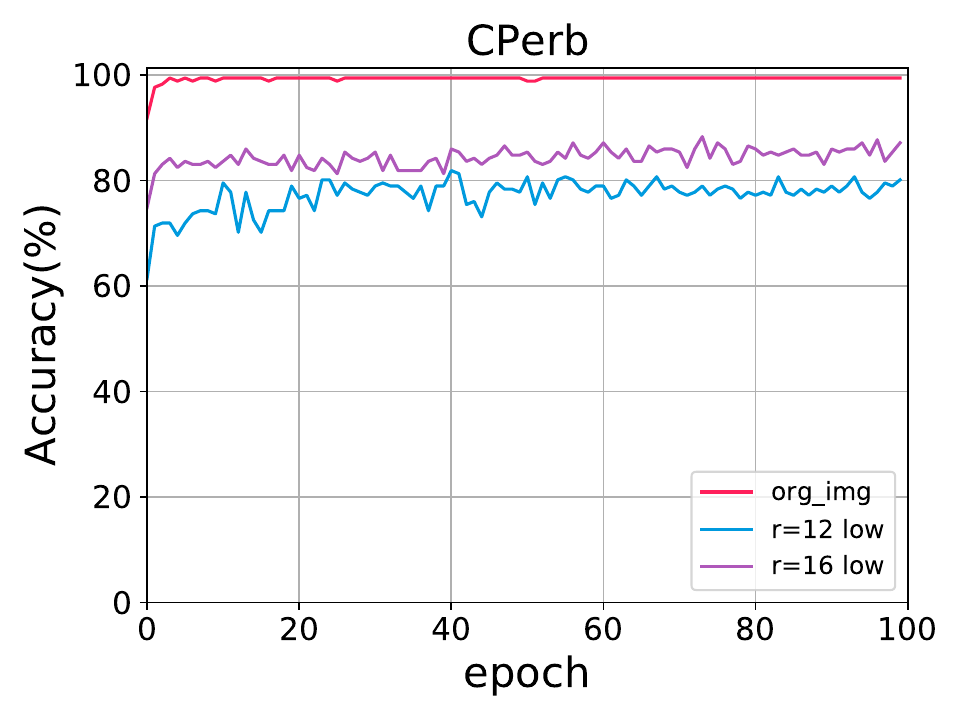}
}
\subfigure[Cartoon]{
  \includegraphics[width=0.21\linewidth]{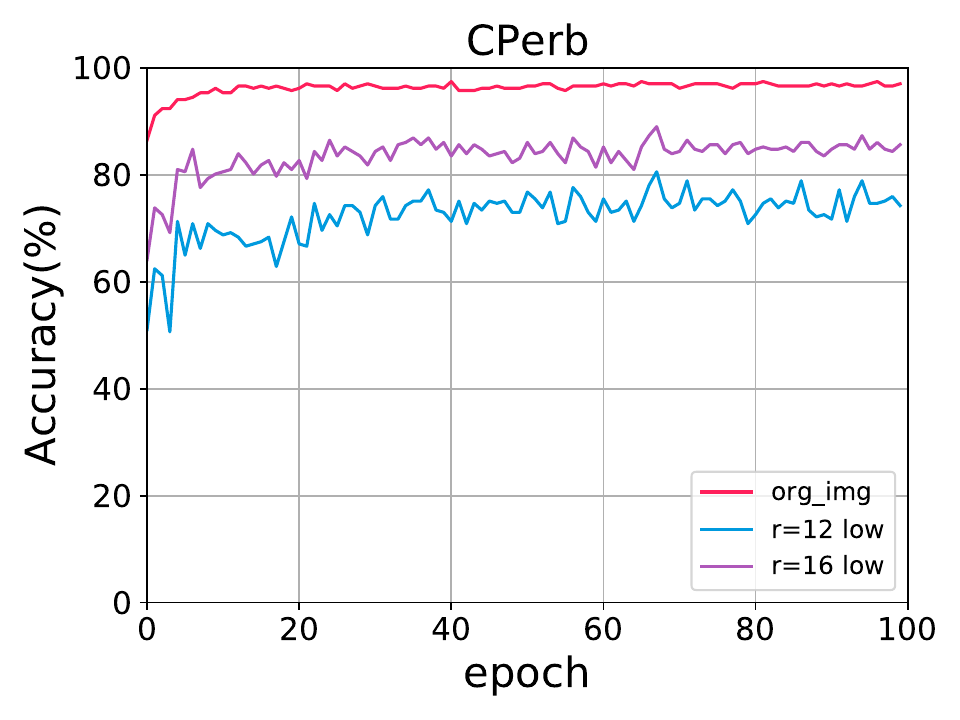}
}
\subfigure[Art]{
  \includegraphics[width=0.21\linewidth]{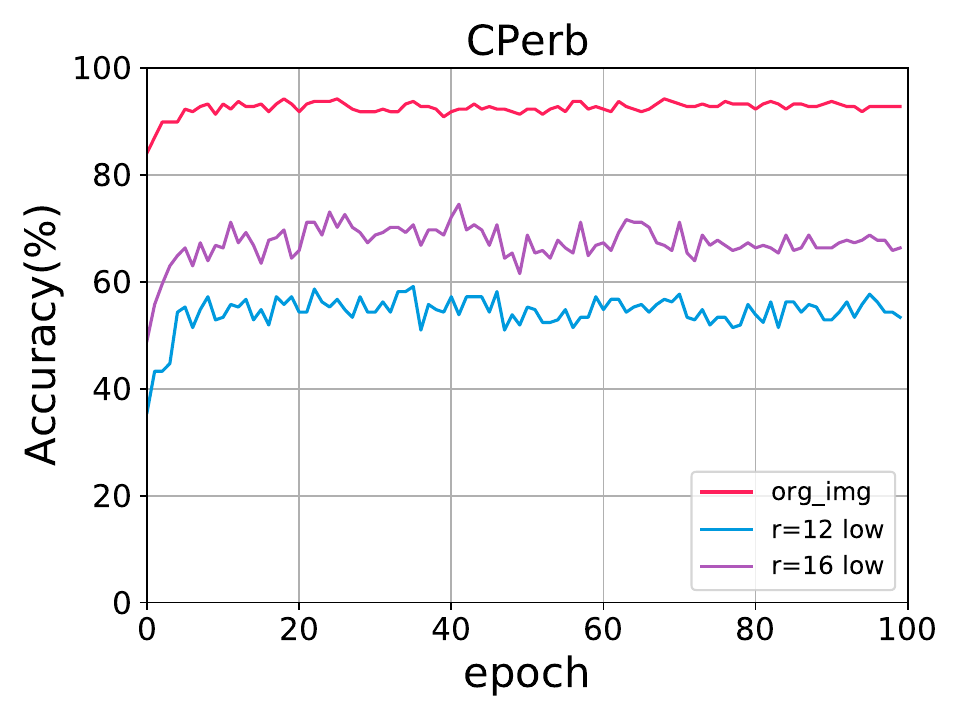}
}
\subfigure[Sketch]{
  \includegraphics[width=0.21\linewidth]{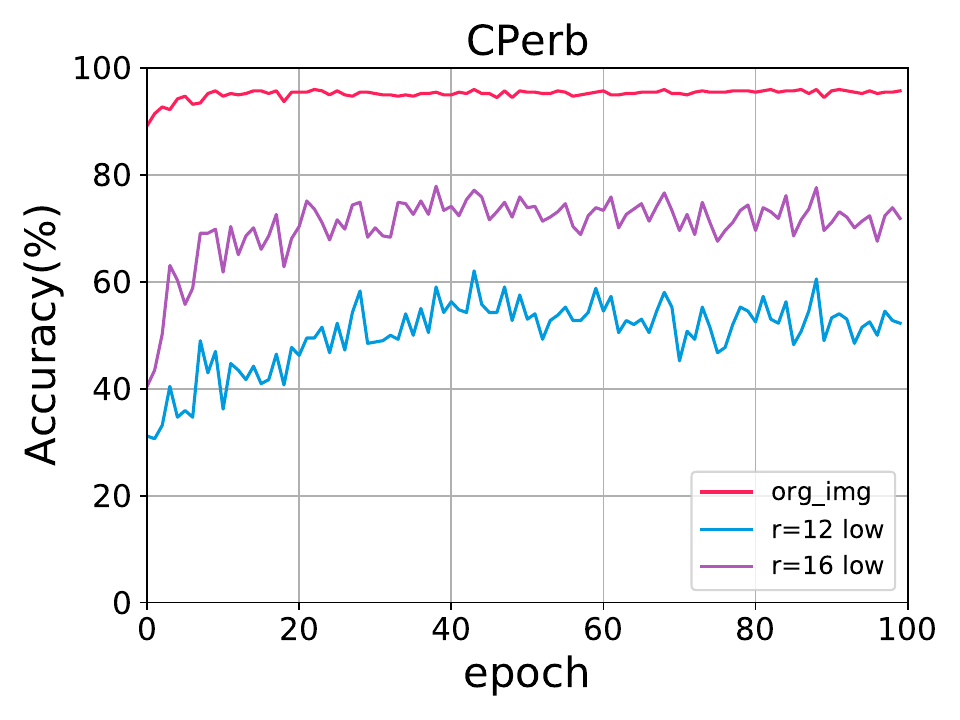}
}
\end{minipage}

\caption{Comparison of accuracy curves between ERM and CPerb for low-frequency component across four distinct domains on the PACS dataset. Each curve represents the validation set results for the respective domain.}
\label{fig9}
\vspace*{-20pt}
\end{figure}

\subsubsection{\textbf{Analysis of the low-frequency component}} \label{sec:EXP-FAHFC}
As demonstrated by~\cite{wang2020high}, the Fourier transform and inverse Fourier transform can be employed to extract the low-frequency components (LFC) and high-frequency components (HFC) from images, which respectively represent domain-invariant and domain-specific features. 
In our investigation, we conduct experiments on the PACS dataset across four distinct domains to verify the efficacy of the CPerb in promoting the acquisition of domain-invariant features.
Specifically, we employ Fourier transform and inverse Fourier transform with varying threshold values (r) to segregate the low-frequency components (LFC).
Subsequently, the LFC is evaluated using the ERM and CPerb frameworks. In Fig.~\ref{fig9}, the \setulcolor{blue_color}\ul{blue} and \setulcolor{purple_color}\ul{purple} line represents the LFC with r set to 12 or 16, respectively. Regarding the domain-invariant features (LFC), the CPerb framework exhibits an average improvement of approximately $18.8\%$ compared to the ERM, thus affirming the efficacy of the multi-route data perturbation in extracting domain-invariant features.

\begin{figure}[t]
\centering
\subfigure[ERM]{
\begin{minipage}[b]{0.42\linewidth}
\centering
\includegraphics[width=\textwidth]{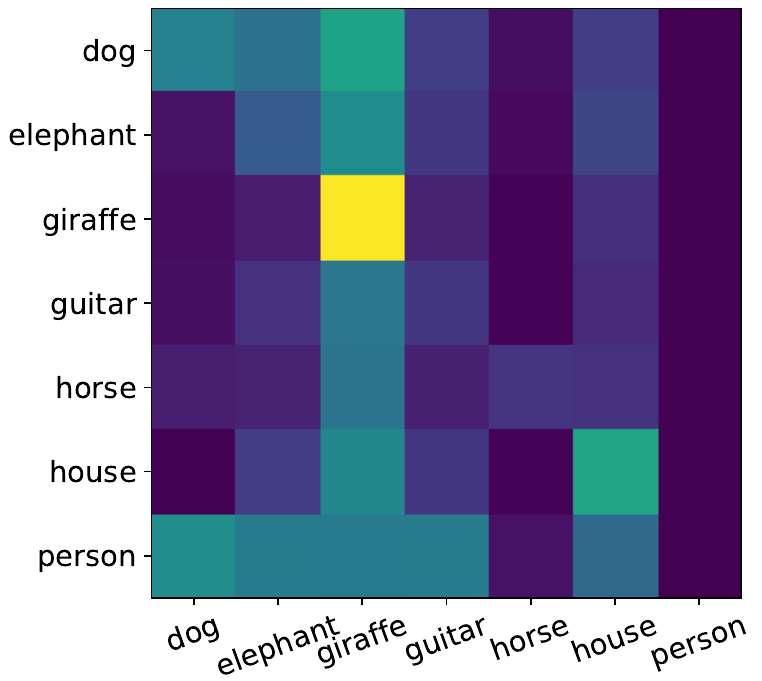}
\end{minipage}
}
\subfigure[CPerb]{
\begin{minipage}[b]{0.48\linewidth}
\centering
\includegraphics[width=\textwidth]{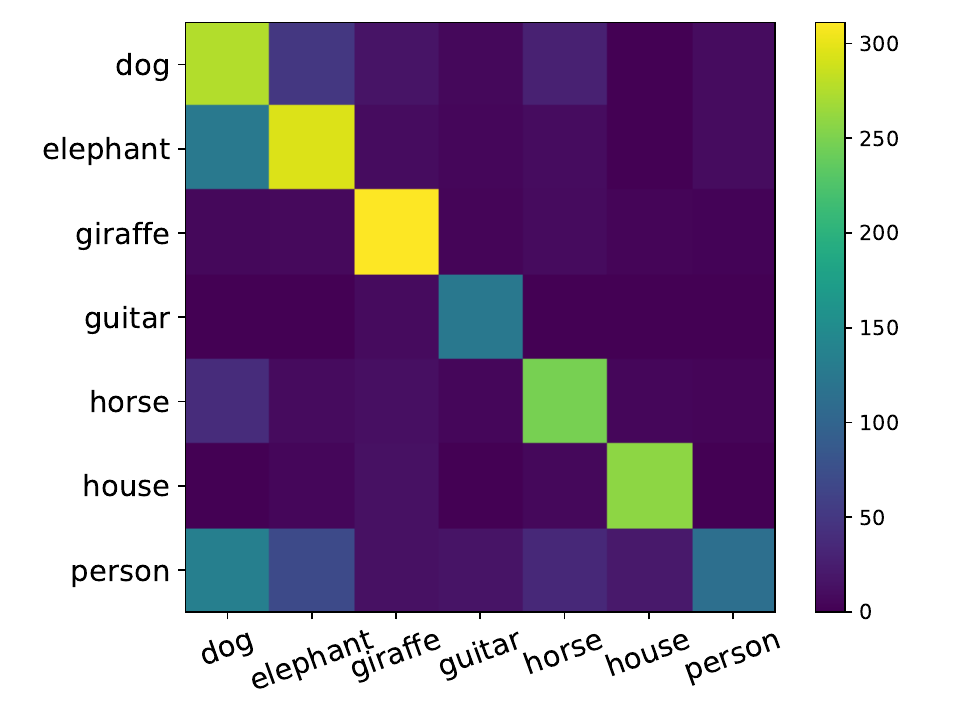}
\end{minipage}
}
\caption{Visualization of confusion matrix: ``sketch'' to ``cartoon'' mapping on the PACS dataset. Figures (a) and (b) illustrate the comparative analysis between the ERM and CPerb. The x-axis of the confusion matrix represents the predicted labels, while the y-axis represents the true labels.}
\label{fig10}
\vspace*{-22pt}
\end{figure}

\subsubsection{\textbf{Visualization for confusion matrix}} \label{sec:EXP-FACM}
In addition to quantitative comparisons highlighting the superiority of the CPerb, we further investigate qualitative distinctions through the visual representation of the confusion matrix. The visualization of prediction outcomes from the ERM and CPerb on the PACS dataset is presented in Fig.~\ref{fig10}, focusing on the domain generalization scenario from ``sketch'' to ``cartoon''. The horizontal axis represents the predicted labels, while the vertical axis corresponds to the true labels. The color gradient along the diagonal of the confusion matrix, shifting towards \setulcolor{yellow_color}\ul{yellow}, signifies improved generalization performance of the model. Notably, even in extreme scenarios with substantial discrepancies between the ``sketch'' and ``cartoon'' domains, the CPerb consistently demonstrates remarkably superior generalization capabilities compared to the ERM, showcasing its exceptional category discrimination prowess.

\subsubsection{\textbf{Visualization for class activation map}} \label{sec:EXP-FACAM}
\begin{figure}[t]
\centering
\includegraphics[width=8.5cm]{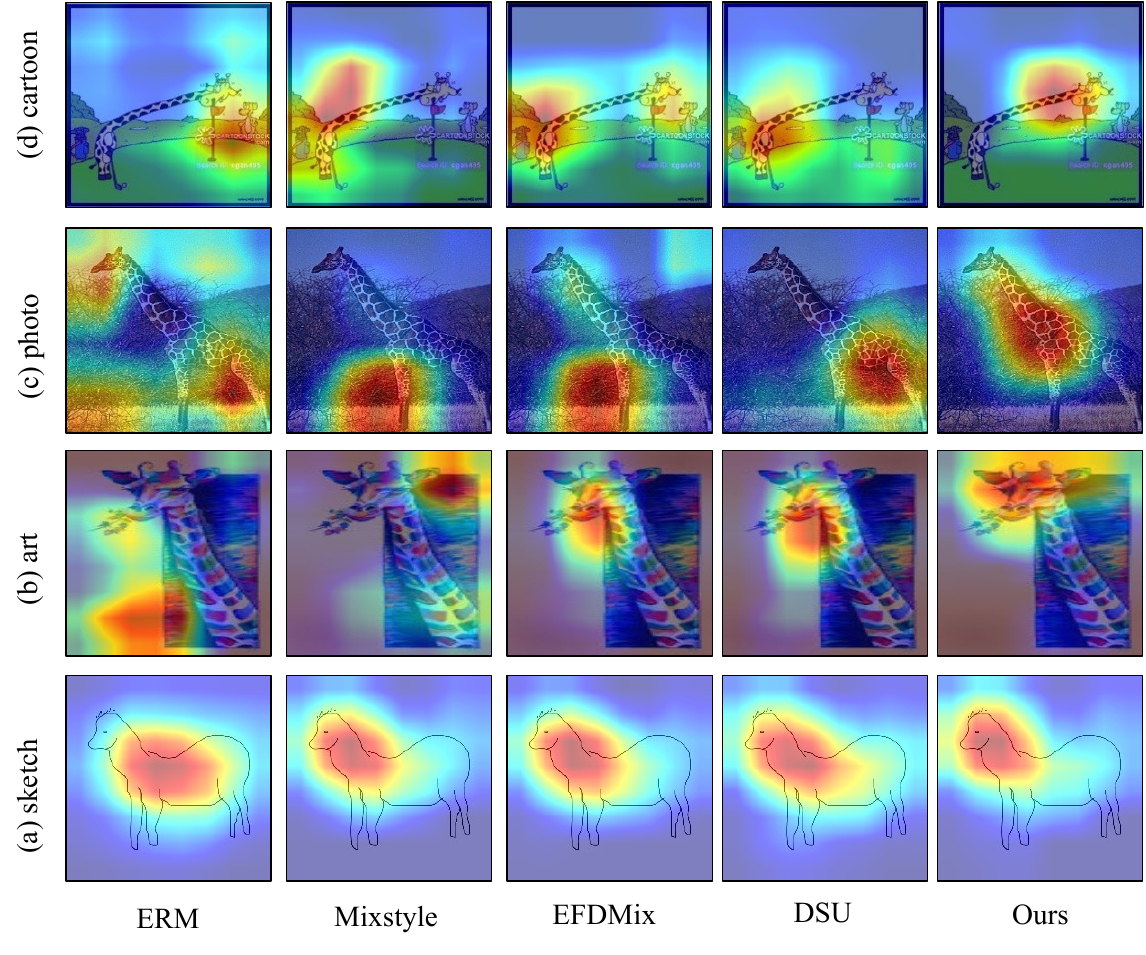}
\caption{Visualization of Class Activation Maps Using GradCAM~\cite{selvaraju2017grad}: The intensity of the red color indicates the level of model attention, with darker regions indicating stronger focus. Each row corresponds to a different domain, with the first three rows representing target domains and the fourth row representing the source domain. Each column denotes a distinct feature perturbation method.}
\label{fig11}
\vspace*{-15pt}
\end{figure}

In the aforementioned analysis, we conduct a comprehensive examination of the CPerb, considering both classification outcomes and generalization feature extraction. In the subsequent exploration, we delve into the image attention regions during the classification process of the CPerb, adopting a data-centric perspective. To visualize these regions, we employ GradCAM~\cite{selvaraju2017grad} on the PACS dataset, designating Sketch as the source domain and the remaining three domains as target domains. The selection of Sketch as the source domain stems from its notable dissimilarity with other domains, thereby presenting a formidable challenge in extracting domain-invariant features. Upon careful examination of the comparative results depicted in Fig.~\ref{fig11}, it becomes apparent that our method places a stronger emphasis on extracting discriminative and domain-invariant features from the images, even in the context of a single-source domain scenario. Indeed, our method exhibits a notable ability to concentrate on essential features such as the head and body of the dog, while effectively filtering out irrelevant domain-specific background stylistic elements. This keen observation provides strong evidence that the incorporation of multi-route perturbation in CPerb significantly aids in extracting domain-invariant features, consequently enhancing the generalization capacity of the model.

\subsubsection{\textbf{Generalization on instance retrieval}}
We conduct experiments in the person retrieval task to validate the effectiveness of the CPerb framework in multitask scenarios. This task aims to match images of the same identity taken by different cameras. Following the methodology described in \cite{zhang2022exact}, we employ the OSNet architecture as the backbone for our experiments. We select the widely-used Market1501~\cite{zheng2015market1501} and GRID~\cite{loy2009grid} datasets, alternating between them as the source and target domains.
As shown in Tab.~\ref{tab08}, the CPerb framework consistently outperforms the current SOTA methods in all evaluation metrics. This is particularly noticeable when using Market1501 as the source domain and GRID as the target domain, where the CPerb framework achieves a significant $3.8\%$ improvement in mean average precision (mAP) compared to the EFDMix~\cite{zhang2022exact}. These results highlight the versatility of the CPerb framework, demonstrating its effectiveness across different backbone architectures and diverse single-source domain generalization tasks.

\begin{table*}[th]
\centering
\caption{Experimental evaluation on ImageNet. ImageNet is source domain, and we test the model in different domains. It is worth noting that for mCE, the smaller value is better; for Top-1 ACC, the larger value is better.}
\label{tab:imagenet}
\setlength{\tabcolsep}{4mm}{
\begin{tabular}{l|c|c|c|c}
\bottomrule
Methods & ImageNet-C (\textcolor{morandired}{mCE$\downarrow$}) & ImageNet-A (Top-1 Acc.$\uparrow$) & Stylized-ImageNet (Top-1 Acc.$\uparrow$) & ImageNet (Top-1 Acc.$\uparrow$) \\
\hline
ResNet-50 & 74.8 & 3.1 & 8.0 & 76.6 \\
AdvProp~\cite{xie2020adversarial} & 69.7 & 4.8 & 10.3 & 77.5 \\
\rowcolor{gray!20} 
CPerb & \textbf{68.2 $\downarrow$\textcolor{morandired}{6.6}} & \textbf{6.4 $\uparrow$\textcolor{morandired}{3.3}} & \textbf{12.3 $\uparrow$\textcolor{morandired}{4.3}} & \textbf{77.8 $\uparrow$\textcolor{morandired}{1.2}} \\
\toprule
\end{tabular}}
\vspace*{-15pt}
\end{table*}

\begin{table}[htbp]
\centering
\caption{Performance assessment on the PACS dataset: leveraging a model founded on the \underline{{\color{tcsvtblue}ViT}} architecture.}
\label{tab:vit}
\begin{tabular}{l|c|c|c|c|c}
\toprule
Methods & Art & Cartoon & Photo & Sketch & Avg \\
\hline
ViT & 71.8 & 69.7 & 46.2 & 47.4 & 58.8 \\
ViT+MixPatch & 73.7 & 76.2 & 50.4 & 57.6 & 64.5 \\
\rowcolor{gray!20} 
ViT+CPerb & \textbf{78.9} & \textbf{83.2} & \textbf{52.3} & \textbf{68.0} & \textbf{70.6} \\
\bottomrule
\end{tabular}
\vspace*{-15pt}
\end{table}

\subsubsection{\textbf{Suitability for ViT}}
ViT has swiftly risen to prominence as a leading model in the field. Leveraging the natural compatibility of our CPerb method with ViT's patch structure at the feature level, we conduct experiments to validate our method on ViT. Given ViT's pre-established patch partitioning, we directly apply MixPatch perturbations to these existing patches. As evidenced in Tab.~\ref{tab:vit}, MixPatch outperforms ViT by $5.7\%$, underscoring its viability within the ViT framework. Furthermore, within the context of our CPerb framework, we achieve an $11.8\%$ enhancement over ViT. This highlights CPerb's efficacy in further augmenting feature diversity on ViT through image-feature cross-level perturbations, thereby bolstering model generalization performance.

\begin{table}[htbp]
\centering
\caption{Enhanced Accuracy of Single-Domain Generalization on DomainNet: Models Trained on Real (R) and Tested on Painting (P), Infograph (I), Clipart (C), Sketch (S), and Quickdraw (Q).}
\label{tab:domainnet}
\setlength{\tabcolsep}{0.9mm}{
\begin{tabular}{l|c|c|c|c|c|c}
\bottomrule
Methods & Painting & Infograph & Clipart & Sketch & Quickdraw & Avg \\
\hline
ERM & 38.1 & 13.3 & 37.9 & 26.3 & 3.4 & 23.8 \\
RandAug~\cite{cubuk2020randaugment} & 41.3 & 13.6 & 41.1 & 30.4 & 5.3 & 26.3 \\
AugMix~\cite{hendrycks2020augmix} & 40.8 & 13.9 & 41.7 & 29.8 & 6.3 & 26.5 \\
ME-ADA~\cite{zhao2020maximum} & 38.0 & 13.1 & 40.3 & 26.8 & 4.5 & 24.5 \\
ACVC~\cite{cugu2022attention} & 41.3 & 12.9 & \textbf{42.8} & 30.9 & 6.6 & 26.9 \\
SimDE~\cite{xu2023simde} & 42.4 & \textbf{14.9} & 41.7 & 31.0 & 6.1 & 27.2 \\
\rowcolor{gray!20} 
CPerb & \textbf{42.9} & 14.8 & 42.1 & \textbf{32.8} & \textbf{8.5} & \textbf{28.2} \\
\toprule
\end{tabular}}
\end{table}

\begin{table}[htbp]
\centering
\caption{Results on Office-Home Dataset for different source domains.}
\label{tab:officehome}
\begin{tabular}{l|c|c|c|c|c}
\bottomrule
Methods & Art & Clipart & Product & Real-World & Avg \\
\hline
ERM & 65.0 & 64.1 & 60.5 & 66.6 & 64.1 \\
MEADA~\cite{zhao2020maximum} & 68.3 & 65.3 & 60.4 & 67.0 & 65.3 \\
SODG-NET~\cite{bele2024learning} & 65.3 & 69.2 & 67.0 & 68.2 & 67.4 \\
\rowcolor{gray!20} 
CPerb & \textbf{68.2} & \textbf{70.4} & \textbf{67.8} & \textbf{68.7} & \textbf{68.8} \\
\toprule
\end{tabular}
\vspace*{-8pt}
\end{table}

\subsubsection{\textbf{Performance on Large-scale Datasets}}
We extend our validation of the effectiveness of the CPerb framework to larger-scale datasets, including DomainNet, OfficeHome, and ImageNet-C, as presented in Tab.~\ref{tab:domainnet}, Tab.~\ref{tab:officehome}, and Tab.~\ref{tab:imagenet}. 
On the DomainNet dataset, our CPerb exhibit a $4.4\%$ improvement over the baseline, surpassing the current state-of-the-art method, SimDE, by $1\%$. 
On the ImageNet dataset, where ImageNet serves as the source domain dataset and ImageNet-C serves as the target domain dataset, we evaluate based on mCE (\textcolor{morandired}{lower is better}) and Acc (\textcolor{morandired}{higher is better}). It is evident that CPerb achieve a $6.6\%$ reduction in mCE on ImageNet-C and a $2\%$ increase in Acc on Stylized-ImageNet. 
On the OfficeHome dataset, CPerb exhibit a $4.7\%$ improvement over the baseline and a $1.4\%$ improvement over the current state-of-the-art method, SODG-NET. 
The aforementioned experimental results convincingly establish the applicability of CPerb on large-scale datasets.

\begin{table}[htbp]
\centering
\caption{Within the CPerb framework, a comparison is made between the MixPatch feature perturbation method with and without shuffling, with the specific shuffling operations detailed in Equation 7 of the main paper.}
\label{tab:shuffle}
\begin{tabular}{l|c|c|c|c|c}
\toprule
Methods & Art & Cartoon & Photo & Sketch & Avg \\
\hline
ERM & 68.6 & 70.2 & 40.5 & 36.0 & 53.8 \\
CPerb (no shuffling) & 81.1 & 80.9 & 59.5 & 61.2 & 70.7 \\
\rowcolor{gray!20} 
CPerb (shuffling) & \textbf{81.7} & \textbf{81.4} & \textbf{62.8} & \textbf{62.6} & \textbf{72.1} \\
\bottomrule
\end{tabular}
\end{table}

\subsubsection{\textbf{The Reasons for Incorporating Shuffling in MixPatch}}
We shuffle within the same channel because different patches exhibit significant variations in statistical information (i.e., different styles). Shuffling enhances the diversity of patch styles, enabling the model to learn more domain-invariant features and consequently improve its generalization to unknown domains. As depicted in Tab.~\ref{tab:shuffle}, CPerb with shuffling exhibits a $1.4\%$ improvement compared to the version without shuffling. 
Furthermore, to tap into potential style information within a reasonable range, we generate new styles post-shuffling using a Gaussian distribution. These styles are randomly sampled within the Gaussian distribution to ensure the introduction of diversified styles without introducing irrelevant noise, thus safeguarding the integrity of the target information.

\section{Conclusion}\label{s-conclusion}
This paper proposes CPerb, a cross-perturbation method, to address the challenges of single-source domain generalization. CPerb employs horizontal and vertical operations to enhance the model's ability to generalize to unknown domains. The horizontal operations alleviate the limited diversity in training data, while the vertical operations facilitate learning domain-invariant features. MixPatch, a novel feature-level perturbation method, further enhances training data diversity. Experimental results validate the effectiveness of CPerb, showing promise in improving generalization performance in single domain generalization tasks. Future work can explore extra methods to enhance CPerb's ability and apply it to other related domains.



%
%

\ifCLASSOPTIONcaptionsoff
  \newpage
\fi

\bibliographystyle{IEEEtran}
\bibliography{sigproc}


\end{document}